\documentclass[letterpaper, 10 pt, conference]{ieeeconf}  

\IEEEoverridecommandlockouts                              

\usepackage{graphics} 
\usepackage{graphicx}
\usepackage{epsfig} 
\usepackage{mathptmx} 
\usepackage{times} 
\usepackage{amsmath} 
\DeclareMathOperator*{\argmax}{arg\,max}

\usepackage{amssymb}  
\usepackage{bm}
\usepackage{algorithm}
\usepackage{algpseudocode}
\usepackage{caption}
\usepackage{subcaption}
\usepackage{physics}
\usepackage{xcolor}
\usepackage{tikz}
\usetikzlibrary{fit,
                positioning}
\usepackage[belowskip=1ex]{subcaption}  

\usetikzlibrary{calc} 
\usetikzlibrary{positioning}
\usepackage{balance}
\usepackage{soul}
\usetikzlibrary{arrows}
\usepackage{cite}
\usepackage{url}

\usepackage{amssymb}
\usepackage{amsfonts}
\usepackage{amsmath}
\usepackage{amsthm}
\usepackage{bm}

\usepackage[normalem]{ulem}                                        
\usepackage{marginnote}
\usepackage[textwidth=10ex,colorinlistoftodos]{todonotes}

\colorlet{mh}{red}
\colorlet{fwu}{red}
\colorlet{ywu}{blue}
\colorlet{kchen}{blue}
\colorlet{lchen}{green}
\colorlet{zbing}{green}
\colorlet{shaddadin}{purple}
\colorlet{iperez}{cyan}
\colorlet{schneider}{magenta}

\usepackage{amssymb}
\usepackage{mathtools}
\usepackage{sansmath}

\mathchardef\mhyphen="2D   

\newcommand{\RNum}[1]{\uppercase\expandafter{\romannumeral #1\relax}}

\newcommand{\bC}     {\mathbf{C}}         
\newcommand{\bD}     {\mathbf{D}}         
\newcommand{\bG}     {\mathbf{G}}         
\newcommand{\bJ}     {\mathbf{J}}         
\newcommand{\bK}     {\mathbf{K}}         
\newcommand{\bM}     {\mathbf{M}}         

\newcommand{\be}{\mathbf{e}}
 
\newcommand{\bq}     {\mathbf{q}}         

\newcommand{\bx}     {\mathbf{x}}         

\newcommand{\btau}{\boldsymbol{\tau}}

\newcommand{\bqdot}  {\dot{\bq}}          

\newcommand{\bqddot} {\ddot{\bq}}         

\title{\LARGE \bf
Contact-aware Shaping and Maintenance of Deformable Linear Objects With Fixtures
}

\author{Kejia Chen$^{1}$, Zhenshan Bing$^{1}$, Fan Wu$^{1}$, Yuan Meng$^{1}$, André Kraft$^{2}$, Sami Haddadin$^{1}$,  Alois Knoll$^{1\dagger}$ 
\thanks{$^{1}$K. Chen, Z. Bing, F. Wu, Y. Meng, S. Haddadin and A. Knoll are with the Department of Informatics, Technical University of Munich, Germany.
        {\tt\small kejia.chen@tum.de}}%
\thanks{$^{2}$ A. Kraft is with BMW AG, Germany.}%
\thanks{$^{\dagger}$The authors acknowledge the financial support by the Bavarian State Ministry for Economic Affairs, Regional Development and Energy (StMWi) for the Lighthouse Initiative KI.FABRIK (Phase 1: Infrastructure as well as the research and development program under grant no. DIK0249). Please note that S. Haddadin has a potential conflict of interest as a shareholder of Franka Emika GmbH.}
\thanks{Corresponding author: Zhenshan Bing.}
}

\renewcommand{\baselinestretch}{0.92}

\begin{document}

\maketitle
\thispagestyle{empty}
\pagestyle{empty}

\begin{abstract}
Studying the manipulation of deformable linear objects has significant practical applications in industry, including car manufacturing, textile production, and electronics automation. 
However, deformable linear object manipulation poses a significant challenge in developing planning and control algorithms, due to the precise and continuous control required to effectively manipulate the deformable nature of these objects.
In this paper, we propose a new framework to control and maintain the shape of deformable linear objects with two robot manipulators utilizing environmental contacts. The framework is composed of a shape planning algorithm which automatically generates appropriate positions to place fixtures, and an object-centered skill engine which includes task and motion planning to control the motion and force of both robots based on the object status. 
The status of the deformable linear object is estimated online utilizing visual as well as force information. 
The framework manages to handle a cable routing task in real-world experiments with two Panda robots and especially achieves contact-aware and flexible clip fixing with challenging fixtures.

\end{abstract}

\section{INTRODUCTION}
\begin{figure}[!t]
    \centering
    \begin{tikzpicture}
    \node[inner sep=0pt] (russell) at (0,0)
    {\includegraphics[width=0.48\textwidth]{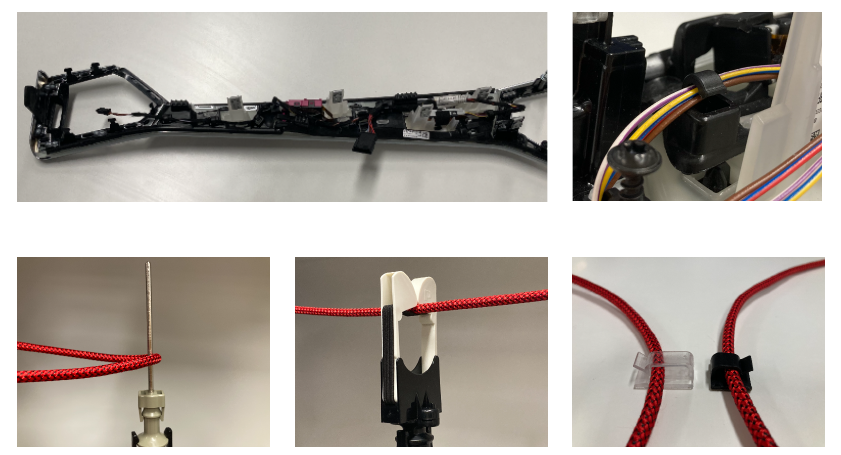}};
    \node at (0,0) {(a)};
    \node at (-2.7,-2.5) {(b)};
    \node at (0,-2.5) {(c)};
    \node at (2.7,-2.5) {(d)};
    \end{tikzpicture}
    \vspace{-1em}
    \caption{Wire harness. (a) Wire harness in the cockpit pre-assembly process of BMW where wires should be installed on a panel (left) using clips (right). (b) Circular fixture. (c) Channel fixture. (d) Clip fixture.}
    \label{fig:use case}
    \vspace{-2em} 
\end{figure}
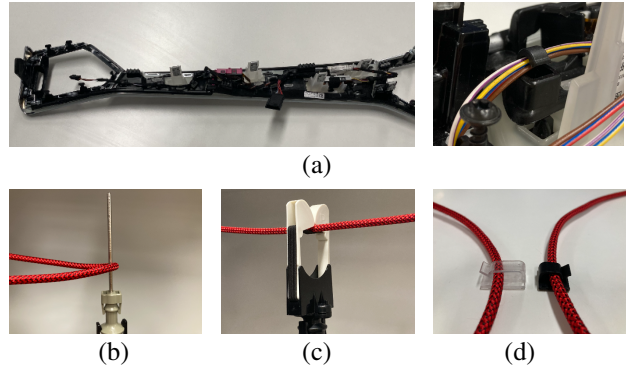

The manipulation of deformable linear objects (DLOs) is a common and yet critical step in various industrial manufacturing processes. One typical example could be the wire harness assembly, where cables need to be installed on a board or panel~\cite{she2021cable}. Despite the fact that most manufacturing processes are now automated, the handling of such deformable cables still heavily relies on manual labor.
Take the cockpit pre-assembly in automotive manufacturing (see Fig.~\ref{fig:use case}(a)) as an example, workers are required to carefully place wires in specified locations, securely fasten them with clips, and repeatedly carry out this process in a continuous manner.

There are several challenges hindering the process of automating the DLO manipulation with robots.
First, it is difficult and time-consuming to model the deformation of an object due to its high dimension, nonlinear behavior, and configuration-dependent properties~\cite{navarro2017fourier,zhu2022challenges}.
Second, typical DLO manipulation mostly relies on  single sensory information, usually the visual perception of the DLO, which is insufficient for estimating the DLO states accurately~\cite{nadon2018multi, sanchez2018robotic, zhu2022challenges}.
Third, DLO manipulation requires multiple robots to work together, along with careful planning and control strategies~\cite{herguedas2019survey, zhu2022challenges}. 
The modeling and state estimation of DLOs, as well as corresponding robot planning, represent significant open problems in the field of DLO manipulation. 


To address the aforementioned challenges, researchers have proposed various frameworks based on different solutions to each sub-problem. Early works focused on achieving a desired manipulation solely with robot contacts~\cite{jia2018manipulating, zhu2018dual, kruse2015collaborative}. These works either did not take environmental contacts into account, or considered such contacts only as obstacles to be avoided in motion planning~\cite{mcconachie2020manipulating}. However, environmental contacts can be assistive, as they provide natural constraints on the object. Therefore, recent works have also aimed to achieve a more complex shape of DLO using contacts from the environment, such as fixtures or clips~\cite{zhu2019robotic, huo2022keypoint, jin2022robotic}.
Nevertheless, the majority of previous works have solely employed circular fixtures or loosely fitted channel fixtures (as shown in Fig.~\ref{fig:use case}(b) and (c)), which result in minimal contact forces with the environment. Although these fixtures can provide temporary constraints on the DLO to control its shape, they are unable to maintain the shape of the DLO, which is often required in production processes. Additionally, most of the works consider only visual perception and robot motion planning, even though the task is contact-rich and has abundant tactile and force information. Moreover, it's important to mention that all of these works began by placing the fixtures in predetermined or random positions, without addressing the issue of finding appropriate fixture positions based on a desired shape.

\begin{figure*}[t!]
    \centering
    \includegraphics[width=0.98\textwidth]{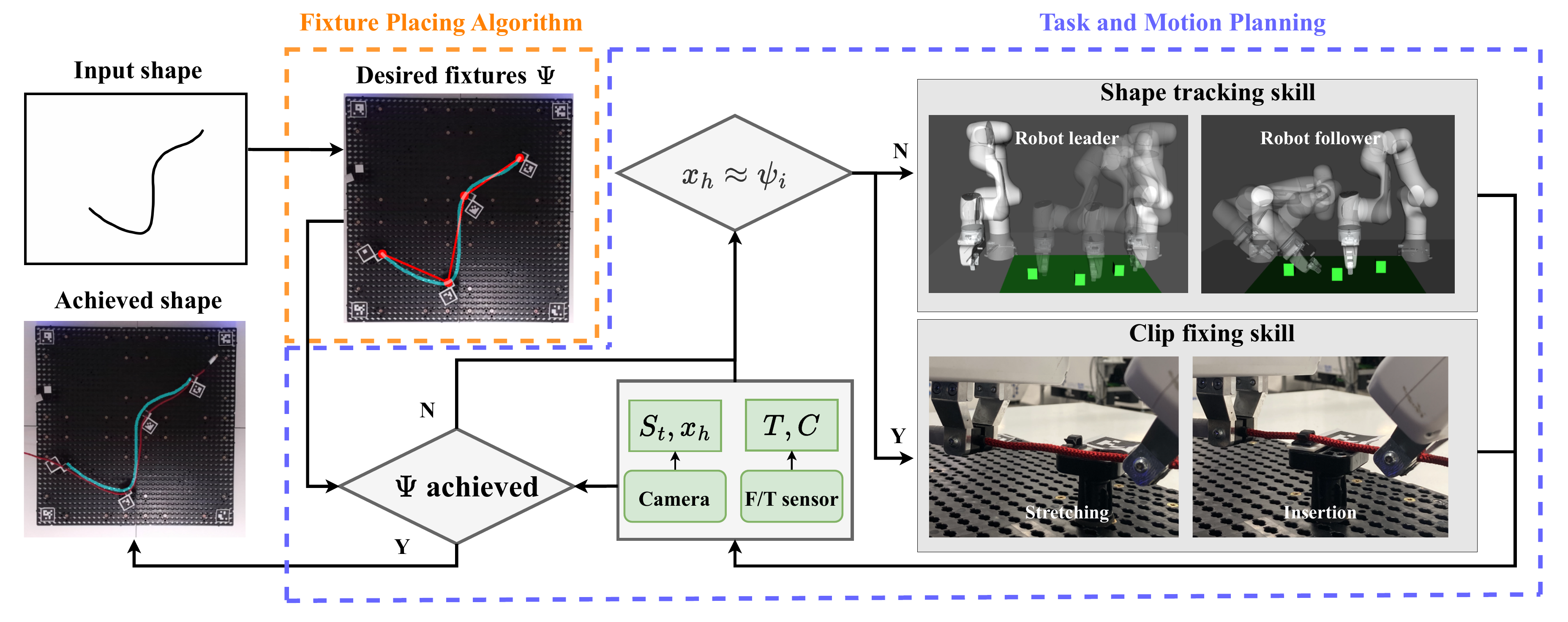}
    \caption{Framework Overview. The framework takes a shape as input, and the fixture placing algorithm (in the dashed orange box) generates appropriate positions to place clips. Based on information from the camera and force sensors, the skill engine for task and motion planning (in the dashed blue box) then executes either the shape tracking skill (top) or the clip fixing skill (bottom) to establish contacts between the DLO and each fixture, and finally achieves the desired shape.}
    \label{fig:overview}
    \vspace{-1.7em} 
\end{figure*}

In this paper, we propose a new framework to manipulate a DLO, which addresses the limitations of prior work by leveraging clip-like fixtures (see Fig.~\ref{fig:use case}(d)), integrating external force sensing, and utilizing an object-centered skill engine. Clip-like fixtures provide more constraints on the object and are thus able to control and at the same time to hold the object's shape. 
Although clip-like fixtures are widely used in industrial manufacturing, they can be challenging to handle even for humans, due to their size and the force required in the fixing process. To address these issues, our framework employs changes in external forces detected by force sensors to achieve adaptive clip fixing and shape control. The contributions of this work are summarized as follows:
\begin{itemize}
    \item A novel fixture placing algorithm is designed to automatically generate positions of fixtures according to the desired shape of a DLO, thereby enabling the DLO to maintain its desired shape and be appropriately manipulated by robots. This algorithm is expected to support wire harness on highly customized shapes in future intelligent manufacturing systems.
    \item We develop an adaptive clip-fixing skill for cautious clipping of DLOs by controlling robots based on contact detection. This skill enables robots to push cables into clips only when the contact is detected, and to cease pushing immediately once the cable fits snugly into the clips, which prevents damage to the objects. By providing a more accurate and careful method for DLO manipulation, this skill possesses the potential to enhance production efficiency and mitigate the occurrence of errors or accidents.
    \item We propose an object-centered framework that combines the aforementioned fixture placing algorithm and clipping skill with collaborative robot motion planning. The new framework integrates visual and external force sensing to estimate the online status of DLOs, based on which the appropriate robot skill is selected for automated and flexible DLO manipulation.  Real-world experiments have validated the effectiveness of our framework in shape control of DLO and demonstrated its advantage over traditional motion planning for achieving a flexible clip fixing process.
\end{itemize}


\begin{figure*}[t!]
     \centering
     \begin{tikzpicture}
     \node[inner sep=0pt] (top) at (-2.5,0)
            {\includegraphics[width=0.45\textwidth]{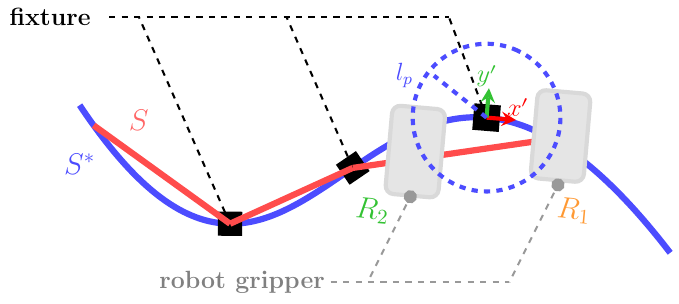}};
     \node[black, font=\bfseries, xshift=0pt, yshift=68pt] at (top.center)  {\large Top View};
     \node[xshift=0pt, yshift=-63pt] at (top.center) {(a)};
     \node[inner sep=0pt] (front) at (5.5,0)
            {\includegraphics[width=0.42\textwidth]{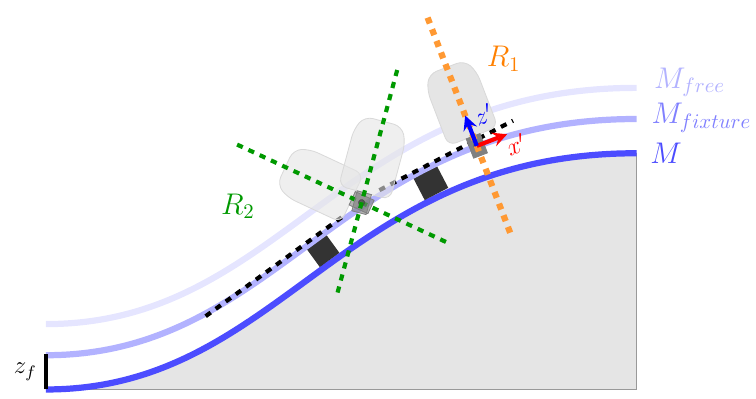}};
     \node[black, font=\bfseries, xshift=0pt, yshift=68pt] at (front.center)  {\large Front View};
     \node[xshift=0pt, yshift=-65pt] at (front.center) {(b)};
     \end{tikzpicture}
     \vspace{-0.7em}
     \caption{Tracking the desired shape. The two gray capsules represent the grippers of two robotic arms. (a) Top view. The selected grasping point on the DLO holds a certain distance (dashed circle) to the current fixture. (b) Front view. The selected end-effector pose of the follower avoids collision with neighboring fixtures as well as with the leader.}
     \label{fig:motion}
\end{figure*}
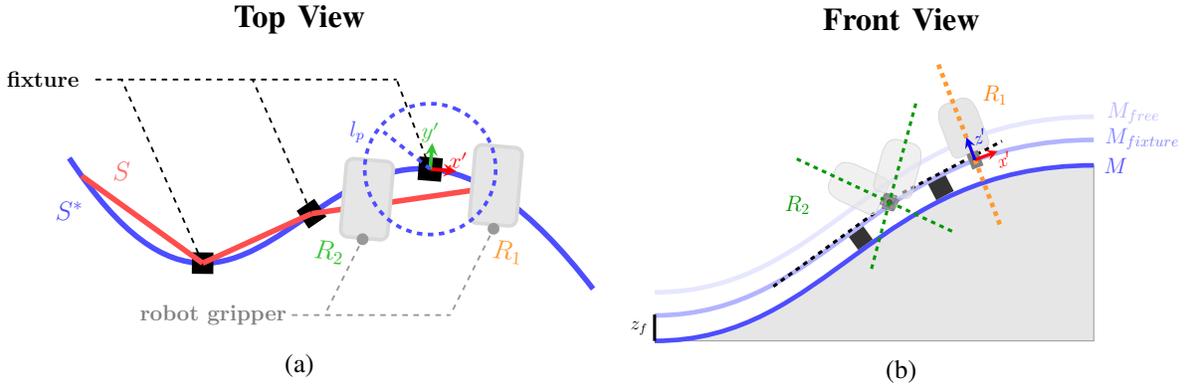
\section{Related Work}
Efforts to control the shape of DLOs with collaborative robot manipulators have traditionally focused only on contacts between the DLO and the robots. Many studies have utilized visual-servoing techniques to establish the relationship between the velocities of the DLO model and the joint velocities of the robot. These techniques rely on different approaches to formulate this relationship, including the use of a dictionary~\cite{jia2018manipulating}, an analytically calculated and updated Jacobian matrix~\cite{zhu2018dual, lagneau2020automatic}, or a neural network to approximate the Jacobian matrix~\cite{yu2022shape}, depending on the visual features chosen for modeling the DLO. Some researchers have taken environmental contacts into account, but only as obstacles to be avoided in robot motion planning. For instance, the planner in~\cite{mcconachie2020manipulating, sintov2020motion} generated robot trajectories that guide the grasped deformable objects through cluttered environments, such as a maze.


  \begin{figure*}[t!]
        \centering 
        \begin{subfigure}[h]{0.8\textwidth} 
        \begin{tikzpicture}
        \node[inner sep=0pt] (curve) at (-1.6,0)
            {\includegraphics[width=0.23\textwidth]{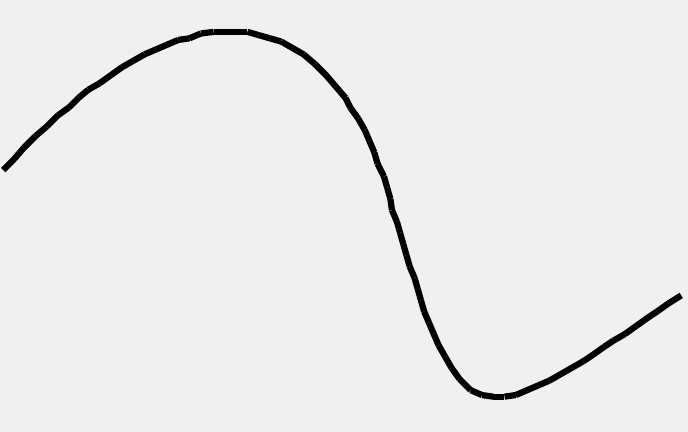}};
        \node[xshift=0pt, yshift=-35pt] at (curve.center) {(a)};
        
        \node[inner sep=0pt] (VE) at (1.8,0)
            {\includegraphics[width=.23\textwidth]{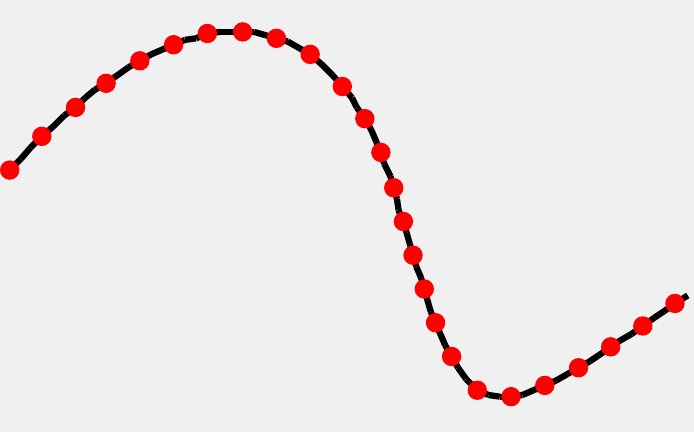}};
        \node[xshift=0pt, yshift=-35pt] at (VE.center) {(b)};
        \node[inner sep=0pt] (flatten) at (7.5,-1.2)
            {\includegraphics[width=0.47\textwidth]{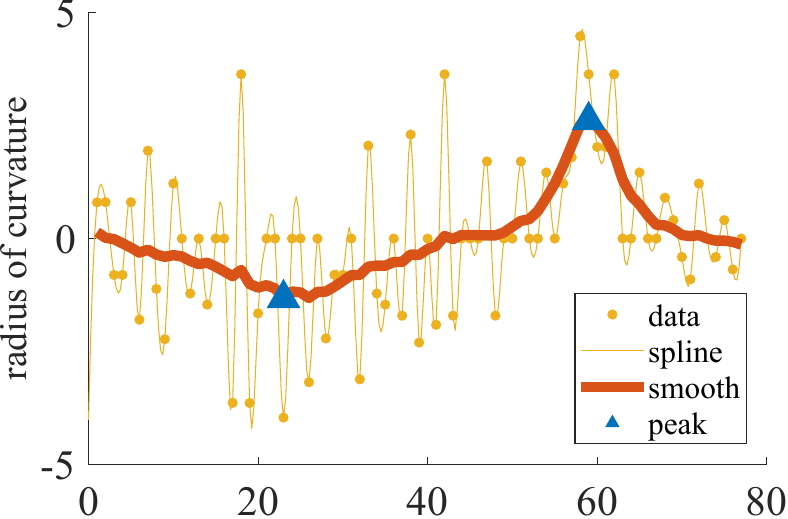}};
        \node[xshift=0pt, yshift=-70pt]at (flatten.center) {(e)};
        \node[inner sep=0pt] (ROC) at (-1.6,-2.5)
            {\includegraphics[width=0.23\textwidth]{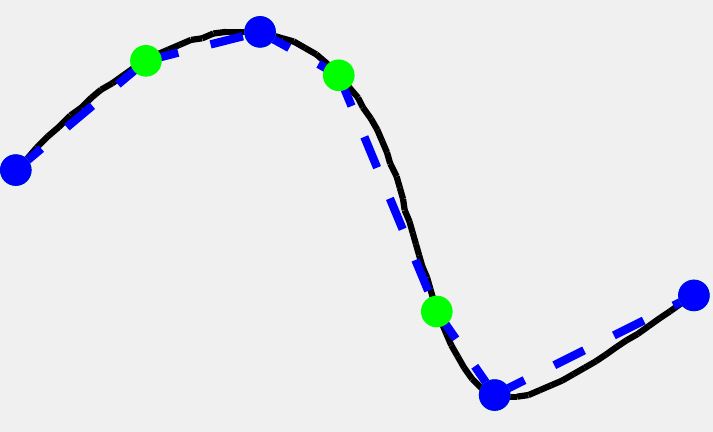}};
        \node[xshift=0pt, yshift=-35pt] at (ROC.center) {(d)};
        \node[inner sep=0pt] (error) at (1.8,-2.5)
            {\includegraphics[width=.23\textwidth]{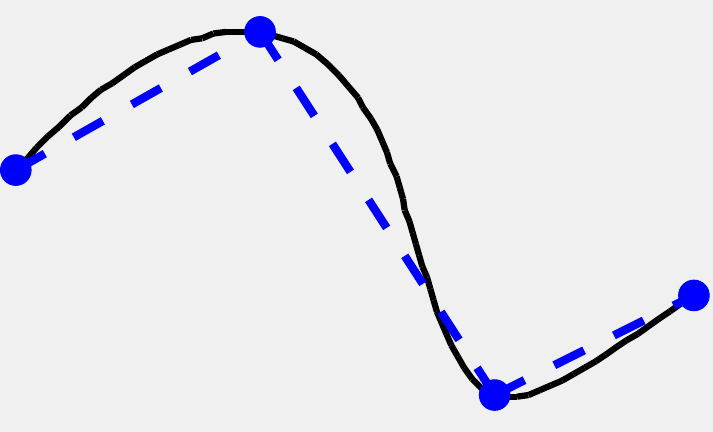}};
        \node[xshift=0pt, yshift=-35pt] at (error.center) {(c)};
        \draw [-stealth][line width =3pt][draw=black] (-0.2,-0.1)--(0.4,-0.1);
        \draw [-stealth][line width =3pt][draw=black] (3.2,-0.1)--(3.8,-0.1);
        \draw [-stealth][line width =3pt][draw=black] (0.4,-2.6)--(-0.2,-2.6);
        \draw [-stealth][line width =3pt][draw=black] (3.8,-2.6)--(3.2,-2.6);
        \end{tikzpicture}
        \end{subfigure}
    \caption{Fixture placing algorithm generating positions of fixtures. (a)-(d) Fixtures on desired shape. The desired shape (a) is firstly represented by vertices and edges (b). Fixtures are then placed at local maximum and minimum ROC (blue markers in c) as well as at maximum error (green markers in d). (e) Analysis of ROC. The raw ROC values are smoothened before locating the local maximum and minimum (blue triangles).}
\label{fig:fixture_position}
\vspace{-1.7em} 
\end{figure*}
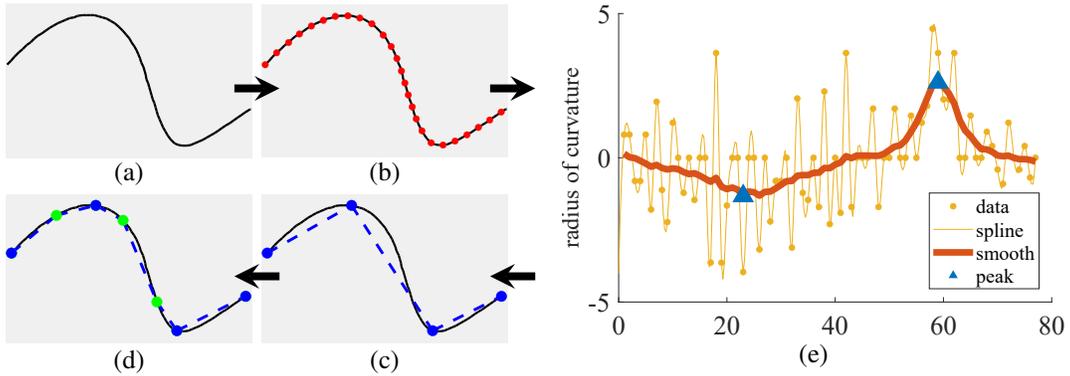

The first framework that controls the shape of DLO using environmental contacts was proposed by Zhu et al.~\cite{zhu2019robotic}. In this framework, a dual-arm robot was used to manipulate DLOs based on circular environmental contacts. Robot motion was planned based on an index that quantified the contact detected by visual sensors, mapping object motions relative to contacts (rotation and sliding) to corresponding robot motions (rotate and pull). Subsequent works~\cite{huo2022keypoint, jin2022robotic, waltersson2022planning} have followed a similar architecture for utilizing environmental contacts, typically consisting of: i) a sensory system that estimates the status of the DLO and detects contacts; ii) a translation system that maps different kinds of DLO movements to corresponding robot motions, forming a group of robot motion primitives; and iii) a primitive-based motion planner.
Huo et al.~\cite{huo2022keypoint} modeled cables as kinematic multi-body systems using a keypoint encoding network and manipulated them using robots under a coarse-to-fine primitive-based planning strategy. Similarly, Jin et al.~\cite{jin2022robotic} took environmental contacts into account in their DLO modeling approach, the spatial representation, which allowed for the generation of intermediate states by the planner, facilitating robotic manipulation.

 
 While previous works on DLO manipulation have differed in terms of DLO modeling and status representation, design of motion primitives and performance indicators, and planning strategies, they have generally relied on visual sensors as the primary source of information. However, DLO manipulation involves a rich amount of force and torque information that is often neglected. In a study on laundry folding, Kruse et al.~\cite{kruse2015collaborative} used a force-feedback controller combined with visual feedback to estimate applied force from external forces as well as desired velocity based on observation of wrinkles on the object. In another study, Suberkrub et al.~\cite{suberkrub2022feel} utilized force-torque information to estimate the status of a DLO modeled as a composition of feature points and edges while keeping the object under tension. A dual-arm robot was then used to manipulate the DLO using nine primitives, each controlling the desired force, position, and orientation of the robot with a hybrid motion-force controller. Despite these advances, external force-torque sensory information is not fully utilized in the manipulation stage.
 
 In our proposed framework, we build upon the general architecture above and integrate both visual and force sensing to achieve effective manipulation of DLOs. Visual sensors are used to assist DLO status estimation and motion planning, while external force information is utilized for clip-fixing, particularly in establishing contacts with environmental fixtures. Additionally, an object-centered task planning system at the high level selects skills for execution based on the difference between the current and desired DLO status.
 
\section{Problem Formulation and Method Overview}
The proposed framework consists of two primary components, as illustrated in Fig.~\ref{fig:overview}: the fixture placing algorithm (in orange box) and a task engine for task and motion planning (in blue box). This section provides a problem statement and an overview of the framework and its individual components.
The framework takes as input a desired shape $S^*$. 
The fixture placing algorithm generates appropriate fixture positions based on radius of curvature (ROC) of points on $S^*$, which allows the DLO to maintain its shape with the help of the fixtures. 
The task and motion planning system then controls the motion and force of both robots based on an online estimation of the DLO status, which is defined as a 4-tuple $(S_t, \mathbf{x}_h, T, C)$.
\begin{itemize}
    \item Real-time shape $S_t$. This is obtained from raw images captured by visual sensors, and can be represented by a sequence of vertices $V = \{v_i\}, i\in\{1, 2,... N\}$ and edges $E= \{e_i\}, i\in\{1, 2,... N-1\}$ following the approach described in Section~\ref{subsec:skill_motion}. $S^*$ can be represented in a similar way as $(V^*, E^*)$.
    \item Position $\mathbf{x}_h$. This is defined as the position of the DLO ``head", i.e., the end that is grasped by the robot. It can be obtained from visual observation or from the forward kinematics of the robot when the head remains grasped by the robot.
    \item Tensity $T \in \{0, 1\}$. When a DLO with low compression strength, such as a rope, is manipulated, it does not offer resistance to deformation in any direction, unless it is under tension~\cite{suberkrub2022feel}. Therefore, tensity is tracked using force sensors as a indicator of deformation resistance. We define $T=1$ if the component of external force $f_{\mathrm{ext}}$ along DLO is above a certain threshold. 
    \item Contact $C\in\{0, 1\}$. This state is obtained from force sensor to indicate DLO's contacts with the environment. Only when the DLO stays taut ($T=1$) are detected contacts with the environment reliable.
\end{itemize}

Based on the observed DLO status, our skill engine selects one from the two designed skill, namely, the shape tracking skill and the clip fixing skill, based on the observed DLO status. The shape tracking skill uses the visual information to generate joint trajectories for both robots, enabling them to track the desired shape $S^*$ and to move to each fixture. In our task, we assume that $S^*$ lies on a surface (two-dimensional manifold) $M$ without any entanglements. This assumption allows us to consider only the projection $\mathbf{x}'\in \mathbb{R}^2$ of a desired end-effector position $\mathbf{x}\in \mathbb{R}^3$ onto two surfaces $M_{\mathrm{fixture}}$ ( the top of fixtures) and $M_{\mathrm{free}}$ (above all fixtures and free of collisions) in trajectory planning of the robot end-effector. $M_{\mathrm{fixture}}$ and $M_{\mathrm{free}}$ are both parallel and holding a different distance to $M$: 
\begin{equation}
    d(M_{\mathrm{free}}, M) > d(M_{\mathrm{fixture}}, M) = z_f\text{,}
\end{equation}
where $z_f$ is the height of fixture. An example of $M$ and $M_{\mathrm{fixture}}$ can be seen in Fig.~\ref{fig:motion}(b).
Since the wiring task is object-centered, we describe the robot motion on surface $M_{\mathrm{fixture}}$ in the DLO frame, with $x'$-axis defined as the tangential direction of DLO and $X'-Y'$ plane as the tangential plane of $M'$ at each point. 
Motion planning for the end-effector of each robot then generates a sequence of desired pose $\mathbf{p}'$:
\begin{equation}
    \mathbf{p}' = ((\mathbf{x}', z'_d), \mathbf{w}')^T\text{,}
\end{equation}
where the distance between $M$ and robots is denoted as $z'_d$, and
the orientation is described by Euler angles $\mathbf{w}' = (w_{x'}, w_{y'}, w_{z'})^T$.


The clip fixing skill uses $T$ and $C$ and applies force that changes in multiple stages, firstly stretching and then pushing the DLO into the fixture. The skill engine applies the shape tracking and clip fixing skills to each fixture in a repetitive manner, until all the intended contacts have been established.

\section{Fixture Planning} \label{sec:placing}

The fixture placing algorithm generates a set of fixture positions $\Psi = 
 \{\mathbf{\psi}\}, \mathbf{\psi}\in\mathbb{R}^2$ on surface $M$ in the DLO frame.
 Taking the continuous desired shape $S^*$ (see Fig.~\ref{fig:fixture_position}(a)) from visual observations as input, we apply a data-driven segmentation method FASTDLO~\cite{caporali2022fastdlo} to convert $S^*$ firstly into a discrete set of vertices and edges $(V^*, E^*)$.
 As is shown in Fig.~\ref{fig:fixture_position}(b), the red points mark the position of $V^*$ and the black edges in-between mark $E^*$.
The segmentation method will be further explained later in Section~\ref{subsec:skill_motion}.

Without loss of generality, we assume that the DLO segments between each consecutive fixture pair are approximately straight, given the forces applied on the DLO by the robots (pulling force from the leader and stretching force from both).
Under such an assumption, fixtures should be placed on surface M at where the desired shape ``bends" the most.
Therefore, we generate fixture position based on the radius of curvature (ROC) $r_i$ of each vertex in $V^*$. 
A spline is established from raw $r_i$ values, and then smoothed by moving average over each window of size $w$, which gives the function $r(v), v \in [1, N]$ (see Fig.~\ref{fig:fixture_position} right column). The first set of fixtures $\Psi = \{\mathbf{\psi}_i\}, i\in\{1, 2,..., L\}$ are then placed on the local maximums and minimums of $r(v)$, splitting $S^*$ into a set of curve segments $S^* = \{s^*_i, s^*_{L+1}\}$ (blue markers in Fig.~\ref{fig:fixture_position}(c)). To maintain the DLO as closely to the desired shape as possible, each fixture's opening direction $\mathbf{u} \in \mathbb{R}^3, \|\mathbf{u}\|=1$ should align with the tangential plane ($X'$-$Y'$) of surface M and be perpendicular to the DLO, resulting in $\mathbf{u} = (0, \pm1, 0)^T$ in the DLO frame. This ensures that the fixtures exert forces in the optimal direction to keep the DLO in its intended shape.

After initial shaping with the planned fixtures, additional fixtures are generated progressively to further reduce the error between the resulting shape of the DLO $S = \{s_i, s_{L+1}\}$ and $S^*$. We define the shape cost function $J_s$ as the sum of area enclosed by each pair of corresponding segments in two shapes:
\begin{equation}
    J_s = \sum_{L+1}\int_0^{l_i} ||s^*_i - s_i|| \text{,}
\end{equation}
where $l_i$ is the length of each segment. On the curve segment $s_k \in S$ contributing the most to $J_s$, an additional fixture (green marker) is placed at the position with the maximum deviation from $s_k^*$. This step is repeated until $J_s$ is below a certain threshold $J_s^*$. 

All the generated fixture positions $\psi_i$ are examined under the global constraint of distance between two neighboring fixtures $d(\psi_i, \psi_j)$. The distance should neither be too narrow for the robot to pass, nor too wide for the object to deform due to its own gravity. This forces fixtures to lie in a space $\mathbb{C}(d)$, which is defined by robot gripper size $l_{g}$ as the lower limit and by maximum allowable sag between consecutive fixtures $J_d$ as the upper limit:
\begin{equation}\label{equ:fixture_global_constriant}
    \mathbb{C}(d) = \{d\in \mathbb{R}^+ | \ d>l_g, f_{\mathrm{deform}}(\kappa, d) < J_d\}\text{,}
\end{equation}
where $\kappa$ defines the stiffness of the DLO, and $f_{\mathrm{deform}}(\kappa, d)$ is the sag of the DLO between two fixtures at a certain distance $d$ to each other. 
The final fixture positions (blue and green markers) and maintained shape (blue dotted line) are shown in Fig.~\ref{fig:fixture_position}(d). The positioning algorithm is summarized in Algorithm \ref{alg:place}.

\begin{algorithm}[t!]
\caption{Place fixture ($S^*$)}\label{alg:place}
\begin{algorithmic}[1]
\State \textbf{Initialize} $\Psi$
\State \{$\psi_i\} =  \argmax_{v}{\| r(v)\|}$  \Comment{Initial shaping}
\If{$\forall \psi \in \Psi\text{,} \ d(\psi_i, \psi) \in \mathbb{C}(d)$}
    \State $\Psi\leftarrow \Psi\cup\{\psi_i \}$
    \State $S \leftarrow \{s_i, s_{L+1}\}$\text{,}\ $S^* \leftarrow \{s^*_i, s^*_{L+1}\}$
\EndIf
\State \textbf{Initialize} $J_s$
\While{$J_s > J^*_s$} \Comment{Additional fixtures}
        \State $k = \argmax_{i} \int_0^{l_i} ||s^*_i - s_i|| $
        \State $\psi_k = \argmax_{v} ||s^*_k(v) - s_k(v)||$
        \State \textbf{repeat step 3-6}
        \State $J_s \leftarrow J_s(\Psi)$
\EndWhile
\State \textbf{return} $\Psi$
\end{algorithmic}
\end{algorithm}
\vspace{-1em} 

\section{DLO Manipulation} \label{sec:manipulation}

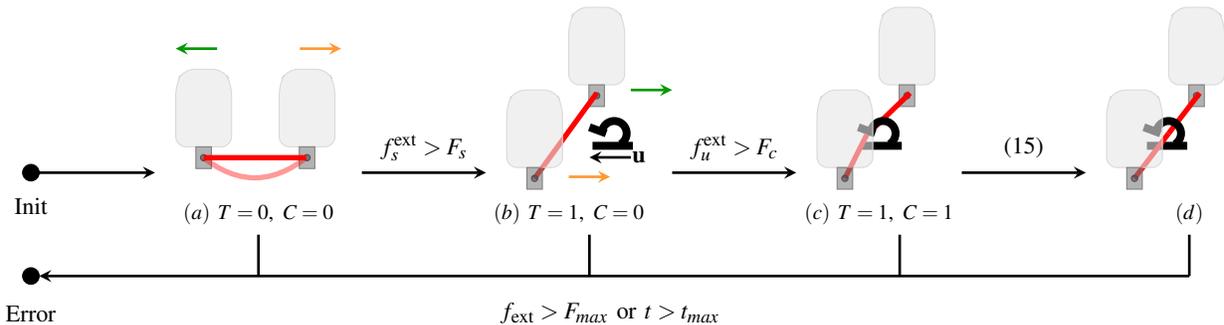
\begin{figure*}[t!]
    \centering
    \begin{tikzpicture}[scale=.55]
        \filldraw (0,0) circle (0.2);
        \node[text=black] at (0, -0.8) {\small Init};
        
        \draw [-stealth][line width =1pt][draw=black] (0,0)--(3, 0);
        
        \filldraw[fill opacity=0.6][rounded corners][draw=gray!30,fill=gray!20, shift ={(3.5 ,-0.5)}, scale =0.75] (0,1.5) -- (1.8,1.5) -- (1.8,3.5) -- (1.55,4)-- (0.25,4) --(0, 3.5) -- cycle;
        \filldraw[fill opacity=0.6][draw=black!50,fill=black!50, shift ={(3.5 ,-0.5)}, scale=0.75] (0.65,0.8) rectangle (1.15, 1.5);
        \filldraw[fill opacity=0.6][draw=black!70,fill=black!70, shift ={(3.5 ,-0.5)}, scale=0.75] (0.9,1.15) circle (.1);
        \draw [-stealth][line width=1pt][draw=orange!80] (6.5,3)--(7.5, 3);
        \filldraw[fill opacity=0.6][rounded corners][draw=gray!30,fill=gray!20, shift ={(6 ,-0.5)}, scale =0.75] (0,1.5) -- (1.8,1.5) -- (1.8,3.5) -- (1.55,4)-- (0.25,4) --(0, 3.5) -- cycle;
        \filldraw[fill opacity=0.6][draw=black!50,fill=black!50, shift ={(6 ,-0.5)}, scale=0.75] (0.65,0.8) rectangle (1.15, 1.5);
        \filldraw[fill opacity=0.6][draw=black!70,fill=black!70, shift ={(6 ,-0.5)}, scale=0.75] (0.9,1.15) circle (.1);
        \draw [-stealth][line width =1pt][draw=black!40!green] (4.5,3)--(3.5, 3);
        \draw[line width=2pt, color=red,  shift ={(3.5 ,-0.5)}, scale=0.75] (0.9,1.15)--(4.2, 1.15);
        \draw[line width=2pt, color=red,  shift ={(3.5 ,-0.5)}, scale=0.75, opacity=0.4] (0.9,1.15) parabola bend (2.55, 0.5) (4.2, 1.15);
        \node[text=black] at (5.5, -1) {\footnotesize $(a)\ T=0, \ C=0$};
        \draw [line width =1pt][draw=black] (5.5, -1.5)--(5.5, -2.5);
        
        \draw [-stealth][line width =1pt][draw=black] (8,0)--(11, 0);
        \node[text=black] at (9.5, 0.6) {\small$f_s^{\mathrm{ext}}>F_s$};
        
        \filldraw[fill opacity=0.6][rounded corners][draw=gray!30,fill=gray!20, shift ={(13 ,1)}, scale =0.75] (0,1.5) -- (1.8,1.5) -- (1.8,3.5) -- (1.55,4)-- (0.25,4) --(0, 3.5) -- cycle;
        \filldraw[fill opacity=0.6][draw=black!50,fill=black!50, shift ={(13 ,1)}, scale=0.75] (0.65,0.8) rectangle (1.15, 1.5);
        \filldraw[fill opacity=0.6][draw=black!70,fill=black!70, shift ={(13 ,1)}, scale=0.75] (0.9,1.15) circle (.1);
        \draw [-stealth][line width =1pt][draw=black!40!green] (14.5,2)--(15.5, 2);
        \draw[line width=2pt, color=red, shift={(11.5 ,-1)}, scale=0.75] (0.9,1.15)--(2.9,3.9);
        \node[text=black] at (13, -1) {\footnotesize $(b)\ T=1,\ C=0$};
        \filldraw[fill opacity=0.6][rounded corners][draw=gray!30,fill=gray!20, shift ={(11.5 ,-1)}, scale =0.75] (0,1.5) -- (1.8,1.5) -- (1.8,3.5) -- (1.55,4)-- (0.25,4) --(0, 3.5) -- cycle;
        \filldraw[fill opacity=0.6][draw=black!50,fill=black!50, shift ={(11.5 ,-1)}, scale=0.75] (0.65,0.8) rectangle (1.15, 1.5);
        \filldraw[fill opacity=0.6][draw=black!70,fill=black!70, shift ={(11.5 ,-1)}, scale=0.75] (0.9,1.15) circle (.1);
        \draw [-stealth][line width=1pt][draw=orange!80] (13,-0.1)--(14,-0.1);
        \draw [line width=3pt,  shift ={(13.5 ,0.5)}, scale =0.15](0,4)--(2.7,3);
        \draw [line width=3pt,  shift ={(13.5 ,0.5)}, scale =0.15] (2,3) arc (180:-60:2);
        \draw [line width=3pt,  shift ={(13.5 ,0.5)}, scale =0.15] (0.5,1)--(7,1);
        \draw [-stealth][line width =1pt][draw=black] (14.5,0.37)--(13.5, 0.37);
         \node[text=black] at (14.7, 0.37) {\small$\textbf{u}$};
        \draw [line width =1pt][draw=black] (13.5, -1.5)--(13.5, -2.5);

        \draw [-stealth][line width =1pt][draw=black] (15.5,0)--(18.5, 0);
        \node[text=black] at (17, 0.6) {\small$f^{\mathrm{ext}}_u>F_c$};
        
        \filldraw[fill opacity=0.6][rounded corners][draw=gray!30,fill=gray!20, shift ={(20.5 ,1)}, scale =0.75] (0,1.5) -- (1.8,1.5) -- (1.8,3.5) -- (1.55,4)-- (0.25,4) --(0, 3.5) -- cycle;
        \filldraw[fill opacity=0.6][draw=black!50,fill=black!50, shift ={(20.5 ,1)}, scale=0.75] (0.65,0.8) rectangle (1.15, 1.5);
        \filldraw[fill opacity=0.6][draw=black!70,fill=black!70, shift ={(20.5, 1)}, scale=0.75] (0.9,1.15) circle (.1);
        \draw[line width=2pt, color=red, shift={(19 ,-1)}, scale=0.75] (1.6, 2.55)--(3,3.9);
        \node[text=black] at (20.5, -1) {\footnotesize $(c)\ T=1,\ C=1$};
        \draw [line width=3pt,  shift={(20 ,0.5)}, scale =0.15](0,4)--(2.7,3);
        \draw [line width=3pt,  shift={(20 ,0.5)}, scale =0.15] (2,3) arc (180:-60:2);
        \draw [line width=3pt,  shift={(20 ,0.5)}, scale =0.15] (0.5,1)--(7,1);
        \draw[line width=2pt, color=red, shift={(19 ,-1)}, scale=0.75] (0.9,1.15)--((1.6, 2.55);
        \filldraw[fill opacity=0.6][rounded corners][draw=gray!30,fill=gray!20, shift ={(19 ,-1)}, scale =0.75] (0,1.5) -- (1.8,1.5) -- (1.8,3.5) -- (1.55,4)-- (0.25,4) --(0, 3.5) -- cycle;
        \filldraw[fill opacity=0.6][draw=black!50,fill=black!50, shift ={(19 ,-1)}, scale=0.75] (0.65,0.8) rectangle (1.15, 1.5);
        \filldraw[fill opacity=0.6][draw=black!70,fill=black!70, shift ={(19 ,-1)}, scale=0.75] (0.9,1.15) circle (.1);
        \draw [line width =1pt][draw=black] (21, -1.5)--(21, -2.5);
        
        \draw [-stealth][line width =1pt][draw=black] (22.5,0)--(25.5, 0);
        \node[text=black] at (24, 0.6) {\small \eqref{eq:lose_contact}};
        
        \filldraw[fill opacity=0.6][rounded corners][draw=gray!30,fill=gray!20, shift ={(27.5 ,1)}, scale =0.75] (0,1.5) -- (1.8,1.5) -- (1.8,3.5) -- (1.55,4)-- (0.25,4) --(0, 3.5) -- cycle;
        \filldraw[fill opacity=0.6][draw=black!50,fill=black!50, shift ={(27.5 ,1)}, scale=0.75] (0.65,0.8) rectangle (1.15, 1.5);
        \filldraw[fill opacity=0.6][draw=black!70,fill=black!70, shift ={(27.5, 1)}, scale=0.75] (0.9,1.15) circle (.1);
        \draw[line width=2pt, color=red, shift={(26 ,-1)}, scale=0.75] (2, 2.6)--(3, 3.9);
        \draw [line width=3pt,  shift={(26.9,0.5)}, scale=0.15](0,4)--(2.7,3);
        \draw [line width=3pt,  shift={(26.9,0.5)}, scale=0.15] (2,3) arc (180:-60:2);
        \draw [line width=3pt,  shift={(26.9,0.5)}, scale=0.15] (0.5,1)--(7,1);
        \draw[line width=2pt, color=red, shift={(26 ,-1)}, scale=0.75] (0.9,1.15)--(2, 2.6);
        \node[text=black] at (28, -1) {\footnotesize $(d)$};
        \filldraw[fill opacity=0.6][rounded corners][draw=gray!30,fill=gray!20, shift ={(26 ,-1)}, scale =0.75] (0,1.5) -- (1.8,1.5) -- (1.8,3.5) -- (1.55,4)-- (0.25,4) --(0, 3.5) -- cycle;
        \filldraw[fill opacity=0.6][draw=black!50,fill=black!50, shift ={(26 ,-1)}, scale=0.75] (0.65,0.8) rectangle (1.15, 1.5);
        \filldraw[fill opacity=0.6][draw=black!70,fill=black!70, shift ={(26 ,-1)}, scale=0.75] (0.9,1.15) circle (.1);
        
        \draw[line width =1pt][draw=black] (28,-1.5)--(28, -2.5);
        \draw[-stealth][line width =1pt][draw=black] (28,-2.5)--(0.2, -2.5);
        \node[text=black] at (14, -3.4) {\small $f_{\mathrm{ext}}>F_{max}$ or $t>t_{max}$};

        \filldraw (0,-2.5) circle (0.2);
        \node[text=black] at (0, -3.4) {\small{Error}};
        
    \end{tikzpicture}
    \caption{Illustration of the clip fixing skill. The two gray capsules represent the grippers of two robotic arms. The skill starts from stretching the DLO to be taut (a) and moves it to establish contact with the fixture (b). Robots then insert the DLO into the clip (c) and stop themselves if the DLO status shows it has a loose fit (d). The entire process operates under the constraints of the maximum allowable external force and a maximum duration. The skill transfers to an error state and stops if either of these constraints is violated.}
\label{fig: fix skill}
\vspace{-1.7em} 
\end{figure*}

The manual cable routing process exhibits a repetitive pattern in desired motion and force at each fixture. In this section, we present our approach of designing two manipulation skills as well as an object-centered high-level task planner to handle this routine.

In both skills, the two robots collaborate in a ``master-slave" manner. The robot leader tracks the desired shape $S^*$ with one end of the DLO grasped by its end-effector throughout the process. The leader maintains its heading to $S^*$ and moves the DLO body from fixture to fixture. Meanwhile, the other robot plans its motion and force application based on the leader's movement. The follower avoids its collision with the leader, and only grasps the DLO at the fixture for clip fixing.

\subsection{Shape Tracking Skill}\label{subsec:skill_motion}

For each fixture $\psi_i$ generated in the previous section, the shape tracking skill controls the motion of both robots to reach a desired pose pair $(\mathbf{p}^i_1,\mathbf{p}^i_2)$ facilitating the upcoming clip fixing skill.   

For the robot leader, the skill aims to track the desired shape $S^*$. Thus, the desired position $\mathbf{x}'_1$ is selected from $V^*$ under a distance constraint to $\psi_i$. The distance $z'_d$ remains $d(M_{\mathrm{free}}, M_{\mathrm{fixture}})$ in the movement to ensure the leader moves in a collision-free way, and is set to zero once it reaches a fixture:
\begin{equation}
    \ z'_d = \{0, \  d(M_{\mathrm{free}}, M_{\mathrm{fixture}})\}\textbf{.}
\end{equation}

For the robot follower, the skill aligns its end-effector with that of the leader:
\begin{equation} \label{equ:follow constraint}
    \begin{aligned}
        w_{x_{2}'} &= w_{x_{1}'}, & w_{z_{2}'} &= w_{z_{1}'}\text{,}
    \end{aligned}
\end{equation}
and grasp the DLO at the fixture. To select a proper grasping position $\mathbf{x}'_2$, the shape of DLO $S_t$ is tracked online by a data-driven method FASTDLO~\cite{caporali2022fastdlo}. FASTDLO effectively segments individual DLO instances from the background, representing each instance as $(V_t, E_t)$. To enhance prediction accuracy, we introduce an additional assumption in FASTDLO that there is only one DLO in the scene. This assumption proves beneficial in handling occlusions caused by the robot bodies during manipulation.
Similar to the robot leader, $\mathbf{x}'_2$ is then selected from $V_t$ under a distance constraint $l_p$. The selection of $\mathbf{x}'_1$ and $\mathbf{x}'_2$ is depicted in Fig.~\ref{fig:motion}(b).

Following the selection of $\mathbf{x}'_2$, the collision avoidance of the robot follower can be formulated as an optimization problem that the optimal orientation should keep the robot follower at the maximum distance to the robot leader and two neighboring fixtures:
\begin{equation} \label{equ:collision}
    \mathbf{w}_2^* = \mathop{\argmax}_{\mathbf{w}'_2} \sum\nolimits_{\mathbf{p}_k \in \{\mathbf{p}_1, \mathbf{\psi}_1, \mathbf{\psi}_2\}} \|\mathbf{p}_2 - \mathbf{p}_k\|\text{.}
\end{equation}
Given constraint~\eqref{equ:follow constraint}, $||\mathbf{p}'_2 - \mathbf{p}_k||$ can be represented by the distance between $Y'-Z'$ planes of the robot follower and each obstacle. The optimization problem can be then simplified as:
\begin{equation} \label{equ:collision_plane}
    {w_{y'_2}}^* = \mathop{\argmax}_{w_{y'_2}} \sum\nolimits_{\mathbf{p}_k \in \{\mathbf{p}_1, \mathbf{\psi}_1, \mathbf{\psi}_2\}} \frac{\mathbf{b}_{k}\cdot \mathbf{n}_k}{\|\mathbf{n_k}\|},
\end{equation}
where  $\mathbf{n_k}$ is the normal vector of each obstacle's $Y'-Z'$ plane, and $\mathbf{b}_{k}$ is the vector pointing from the follower end-effector to an arbitrary point on that plane. An example of different orientation option of the robot follower is shown in Fig.~\ref{fig:motion}(b). After $\mathbf{p}_2$ is selected for the end-effector, we plan the motion of the robot follower in joint space with RRT* algorithm to avoid possible collision between two robot arms in the process of motion.

\subsection{Clip Fixing Skill} \label{subsec:skill_fixing}

After both robots reach appropriate poses relative to $\psi_i$, the follower will also grasp the DLO. The clip fixing skill then controls both robots to apply forces on the grasped segment of DLO to push it into the clip. Taking $T$ and $C$ status of DLO into account, the skill can achieve an adaptive and contact-aware fixing process so that once the DLO is fitted into the clip, it realizes the success and the stops the robots from applying forces or moving further.

We formulate the fixing skill using an adaptive force impedance controller \cite{Yang2011HumanLike}\cite{johannsmeier2019framework}. Each robot arm is assumed to follow the standard rigid robot dynamics
\begin{equation}
    \bM(\bq)\bqddot + \bC(\bq,\bqdot)\bqdot + \bG(\bq) = \btau_u + \btau_\mathrm{ext},
\end{equation}
where $\bM(\bq)$ is the mass matrix, $\bC(\bq,\bqdot)$ denotes the Coriolis and centrifugal matrix, and $\bG(\bq)$ represents the gravity vector, respectively. Their corresponding expressions in Cartesian space are $\bM_c(\bq),\bC_c(\bq,\bqdot)$ and $\bG_c(\bq)$.
The applied joint torque is $\btau_u$ and $\btau_\mathrm{ext}$ denotes external torque.

The adaptive impedance control law is defined as:
\begin{equation}
    \begin{aligned}
    \btau_u(t) = & \mathbf{J}( \mathbf{q})^\mathsf{T} [ -\mathbf{f}^d(t) - \mathbf{K}_c(t)\mathbf{e} - \mathbf{D}_c \dot{\be} \\
                 & + \bM_c(\bq) \ddot{ \bx }^d + \bC_c(\bq,\bqdot)\dot{\bx}^d + \bG_c(\bq)  ], 
\end{aligned}
\end{equation}
where $\mathbf{e}=\mathbf{x^*} - \mathbf{x}$ and $\mathbf{\dot{e}}=\mathbf{\dot{x}^*} - \mathbf{\dot{x}}$ are the position and velocity error respectively. 
$\bK_c(t)$ and $\bD_c$ are stiffness and damping matrices in Cartesian space. $\bJ(\bq)$ is the robot Jacobian.

The clip fixing skill is defined as a directed transition graph of manipulation primitives (MPs). A single MP consists of a desired linear velocity $\dot{\mathbf{x}}^d \in \mathbb{R}^3$ and feedforward force $\mathbf{f}^d \in \mathbb{R}^3$. Taking a fixture with opening  $\mathbf{u}$ in $+y'$ direction ($\mathbf{u} = (0, 1, 0)^T$) as an example, the fixing skill graph is depicted in Fig.~\ref{fig: fix skill}. Considering the rich contact between the DLO and fixtures in the fix skill, all the transitions between MPs are triggered by changes in $T$ and $C$ status. 

\textbf{Pre-contact stretching} Because of the movement in Section~\ref{subsec:skill_motion}, the DLO is usually not under tension by default, so initially $T=0$. To measure the contact force with environmental fixtures accurately, the object is firstly stretched by both robots in $x'$ axis to be taut:
\begin{equation}
\begin{aligned}
    \dot{\mathbf{x}}_1^d &= [0,\  0, \ 0]^T, & 
     \mathbf{f}_1^d &= [f_s, \ 0,\ 0]^T \text{,}
\end{aligned}
\end{equation}
where $f_s$ is the desired magnitude of stretching force. The robot follower in this MP will move and apply forces in an opposite direction to the leader:
\begin{equation}
\begin{aligned}
    \dot{\mathbf{x}}_2^d &= -\dot{\mathbf{x}}_1^d, &  \mathbf{f}_2^d &= -\mathbf{f}_1^d\text{.}
\end{aligned}
\end{equation}
It is worth noting that the stretching force remains for the whole skill to ensure the DLO is under tension for accurate force sensing.

\textbf{Contact establishment} After the stretching condition is satisfied, the taut DLO is then moved in $-\mathbf{u}$ direction:
\begin{equation}
\begin{aligned}
    \dot{\mathbf{x}}_1^d &= -\mathbf{u}, & 
     \mathbf{f}_1^d &= [f_s, \ 0,\ 0]^T,
\end{aligned}
\end{equation}
until it has contact with the fixture. We update the contact status $C$ by detecting the projection of the external force in the moving direction. A contact happens if the magnitude of the projection $f^{\mathrm{ext}}_u$ rises above threshold $F_c$. The robot follower takes the same motion and forces as the robot leader at this stage. The position of the contact point is memorized as $\mathbf{x}_c$.

\textbf{Push in} Once the contact happens, the DLO is pushed into the clip:
\begin{equation}
\begin{aligned}
    \dot{\mathbf{x}}_1^d &= [0, \ 0,\ 0]^T, & 
     \mathbf{f}_1^d &= [f_s, \ 0,\ 0]^T + f_p \cdot (-\mathbf{u}), 
\end{aligned}
\end{equation}
until the DLO moves further than $\mathbf{x}_c$ in $-\mathbf{u}$ direction and loses contact with the fixture:
\begin{equation}
    \begin{aligned}
       \mathbf{x}_c \cdot (-\mathbf{u}) &< \mathbf{x}_t \cdot (-\mathbf{u}), & f^{\mathrm{ext}}_u < F_c.
    \end{aligned}
    \label{eq:lose_contact}
\end{equation}
 The robot follower takes the same motion and forces as the robot leader.

 \subsection{Skill-based planning}
The task planner at the high level selects from the two manipulation skills to establish contact with each fixture following the logic presented in Fig.~\ref{fig:overview}. The selection strategy depends on current DLO position $x_h$. The tracing skill is employed until both robots reach appropriate poses close to $\psi_i$. After that, the follower will also grasp the DLO and the clip fixing skill is activated. This process is repeated until all fixtures generated in Section~\ref{sec:placing} are achieved.
\section{Experiments} \label{sec:experiments}
\begin{figure}[t!]
    \centering
    \begin{tikzpicture}
    \node[inner sep=0pt] (setup) at (0,0) {\includegraphics[width=0.45\textwidth]{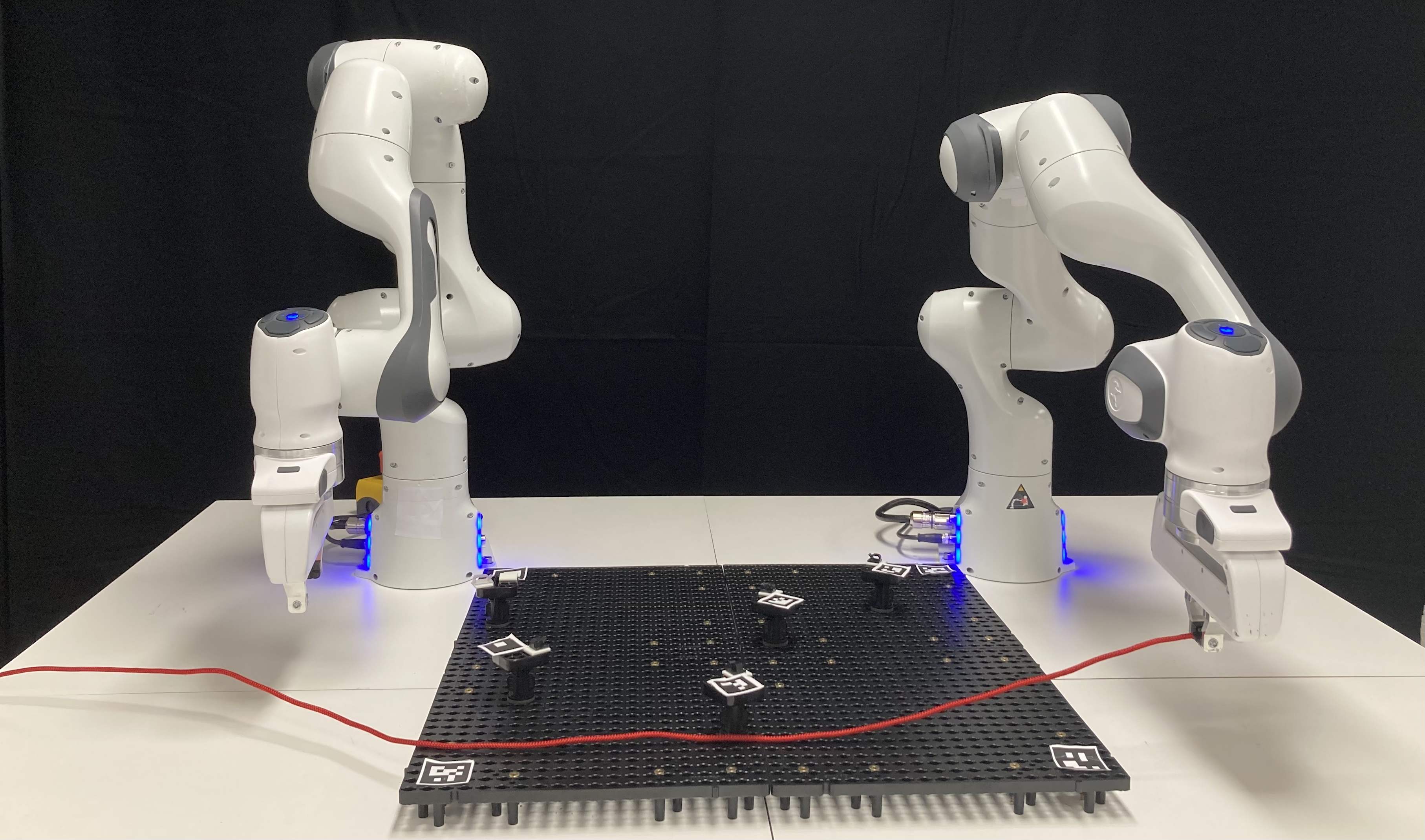}};
    \node[inner sep=0pt, draw=white, thick=5cm] (clip) at (1.5,-1.8) {\includegraphics[width=0.08\textwidth]{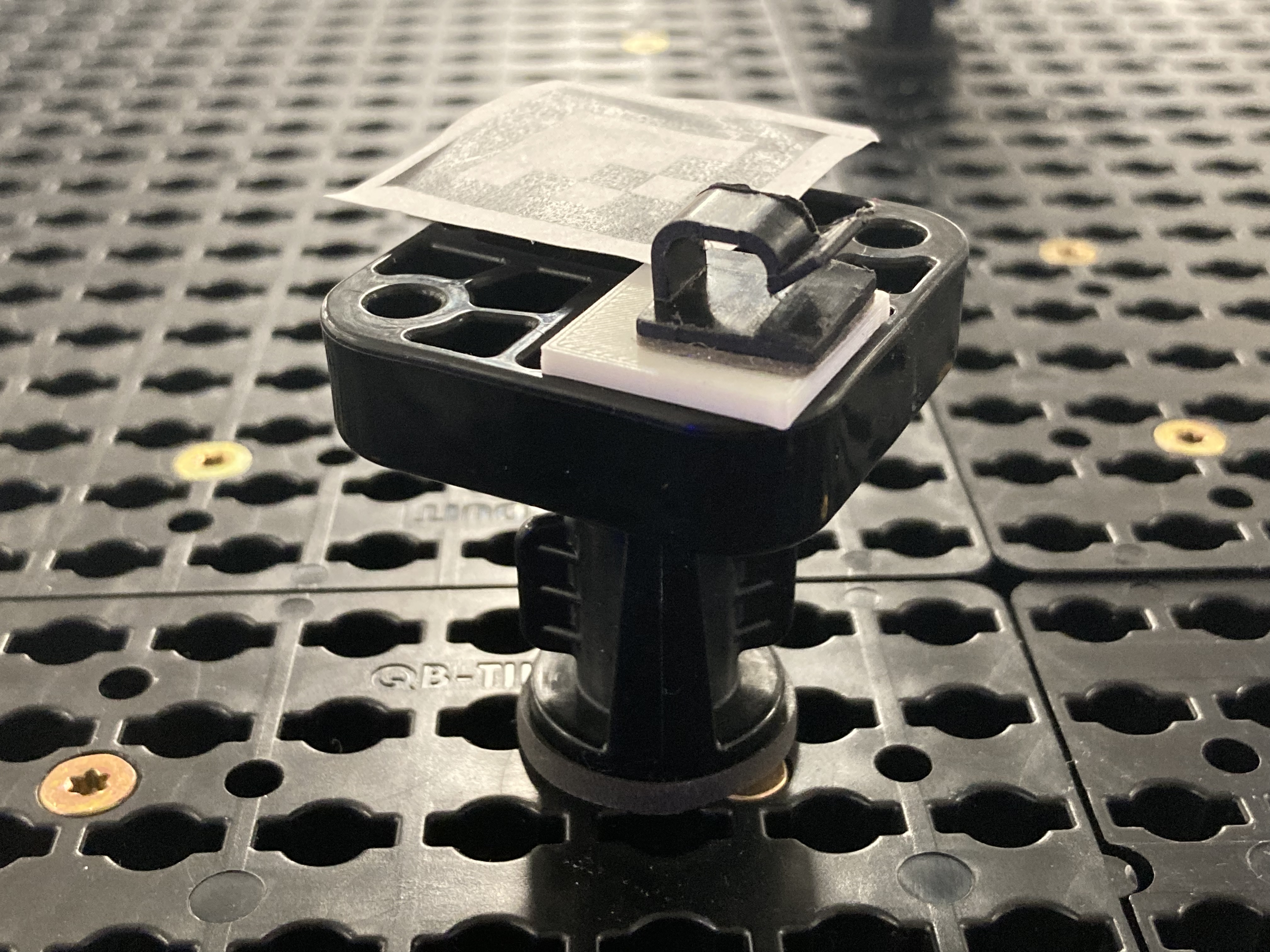}};
    \draw[dashed][line width =1pt][draw=white] (0.1,-1.5)--(clip.north west);
    \draw[dashed][line width =1pt][draw=white] (0.1,-1.5)--(clip.south west);

    \draw[dashed][draw=black!40!green][line width=1pt](-1.5,0.5)--(-1,1);
    \node[text=black!40!green] at (-0.7, 1) {$R_2$};
    \draw[dashed][draw=orange][line width=1pt](1.5,0.5)--(1,1);
    \node[text=orange] at (0.7,1) {$R_1$};
    \draw[dashed][draw=red][line width=1pt](-3,-2)--(-2.6,-1.6);
    \node[text=red] at (-3.3,-2.2) {$DLO$};
    \node[text=white] at (1.3,-2.1) {$\psi_2$};
    \end{tikzpicture}
    \caption{Setup of real-world experiment.}
    \label{fig:exp_setup}
    \vspace{-1.7em} 
\end{figure}
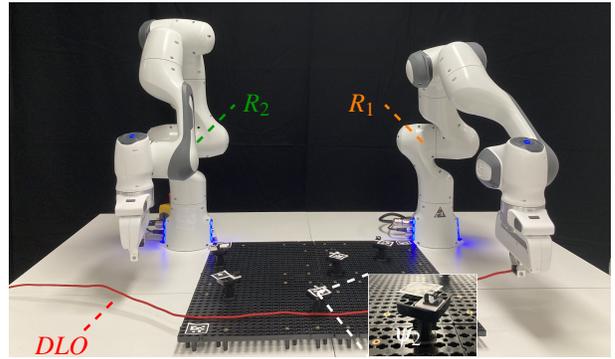

\begin{figure*}[t!]
    \centering
    \begin{subfigure}[b]{0.48\textwidth}
        \centering
        \includegraphics[width=0.54\linewidth]{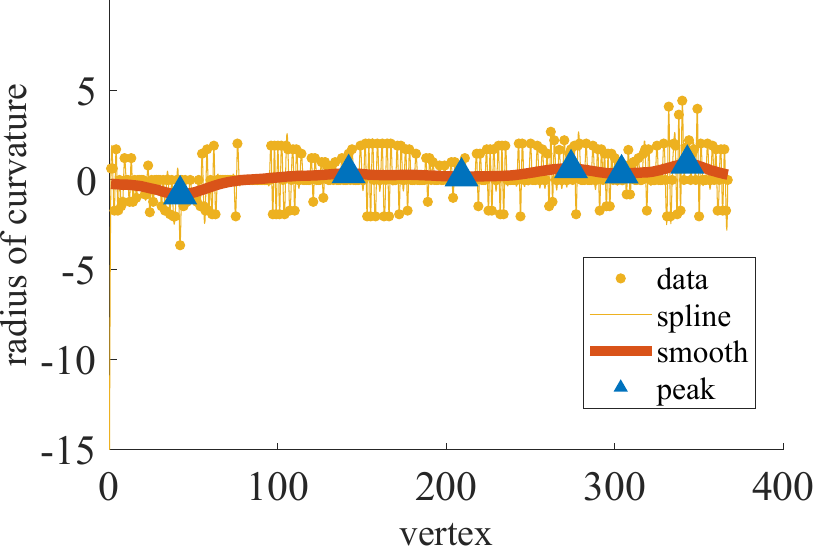}%
        \hfill
        \includegraphics[width=0.45\linewidth]{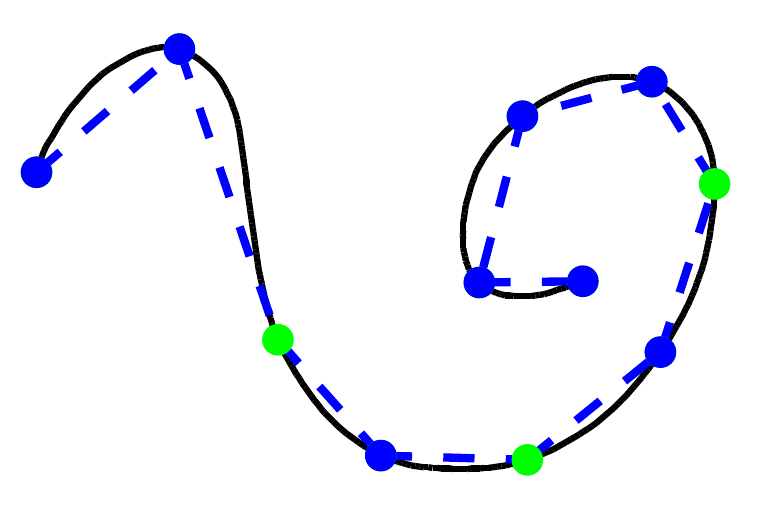}
        \caption{The first desired shape.}
    \end{subfigure}
    \hfill
    \begin{subfigure}[b]{0.48\textwidth}
        \centering
        \includegraphics[width=0.54\linewidth]{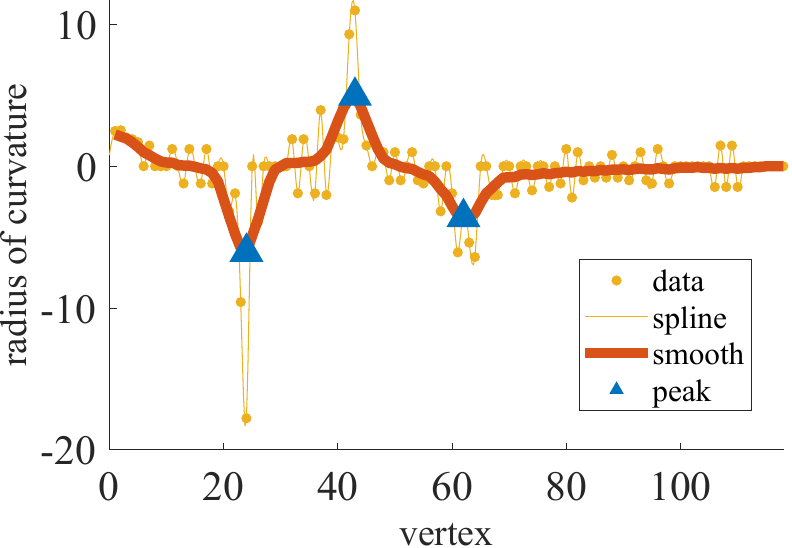}%
        \hfill
        \includegraphics[width=0.4\linewidth]{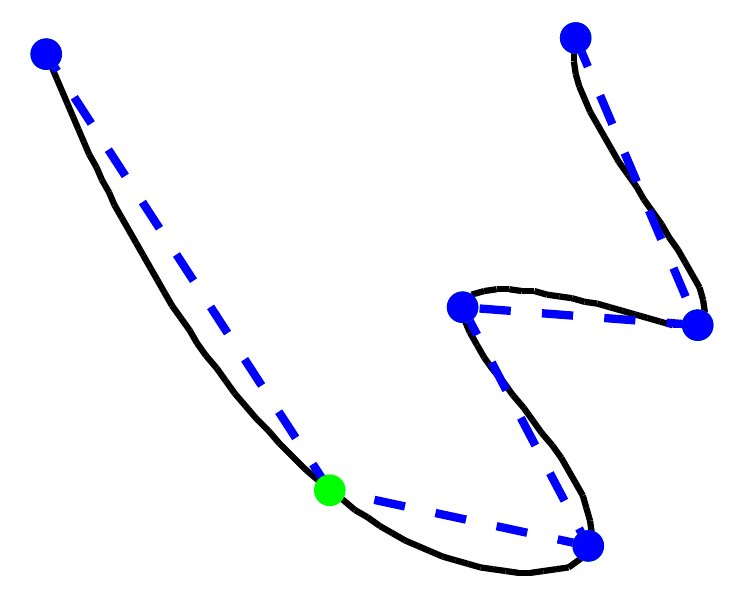}
        \caption{The second desired shape.}
    \end{subfigure}
    \caption{Fixture positions generated by placing algorithm on two desired shapes and corresponding ROCs.}
    \label{fig:exp_placing}
    \vspace{-1.7em}
\end{figure*}

In this section, we present real-world DLO manipulation experiments and evaluate main components of our framework. We use two 7 DOF Franka Emika Panda robots for the experiments, both of which are equipped with joint torque sensors and provide 6-axis force torque estimation at the end-effectors. Since the clips in our task are small, we mount each clip on an additional platform to form a fixture (see Fig.~\ref{fig:exp_setup}). Fixtures mounted on a harness board then define the surface $M$, which is a plane in this case. The positions and orientations of fixtures are estimated from markers by an Azure Kinect camera on the top. To improve the stability of the joint F/T sensor measurements, we further constrain the orientation of robot leader end-effector to be always normal to $M$. In this case, $w_{x'1} = w_{y'1} = 0$ and $w_{z'1}$ alone defines the orientation of robot leader.

\subsection{Fixture Positioning}


Since our robots have grippers that are rather big relative to clips, the workspace is limited according to~\eqref{equ:fixture_global_constriant} and only a few fixtures are allowed to be placed. To show the capability of our placing algorithm to handle more complex shapes, here we run the test with $l'_g$ set as only half of the real gripper size $l'_g = 0.5\cdot l_g$. The deformation function $f_{\mathrm{deform}}$ is obtained by measuring the deformation of the DLO with both ends held by robots. 

Two examples of placing results are presented in Fig.~\ref{fig:exp_placing}. The size of smoothing windows $w$ are chosen depending on the level of noise in the desired shape. For hand-drawn trajectories with small fluctuations, smoothing operations are applied for multiple times. The blue markers in Fig.~\ref{fig:exp_placing} represent fixtures placed at local maximum and minimum of ROC and the green markers represent fixture positions generated recursively to reduce $J_s$.

\begin{figure}[!t]
    \centering


    \begin{tikzpicture}
        \node[inner sep=0pt] (test_eval) at (0,0) {\includegraphics[width=0.48\textwidth]{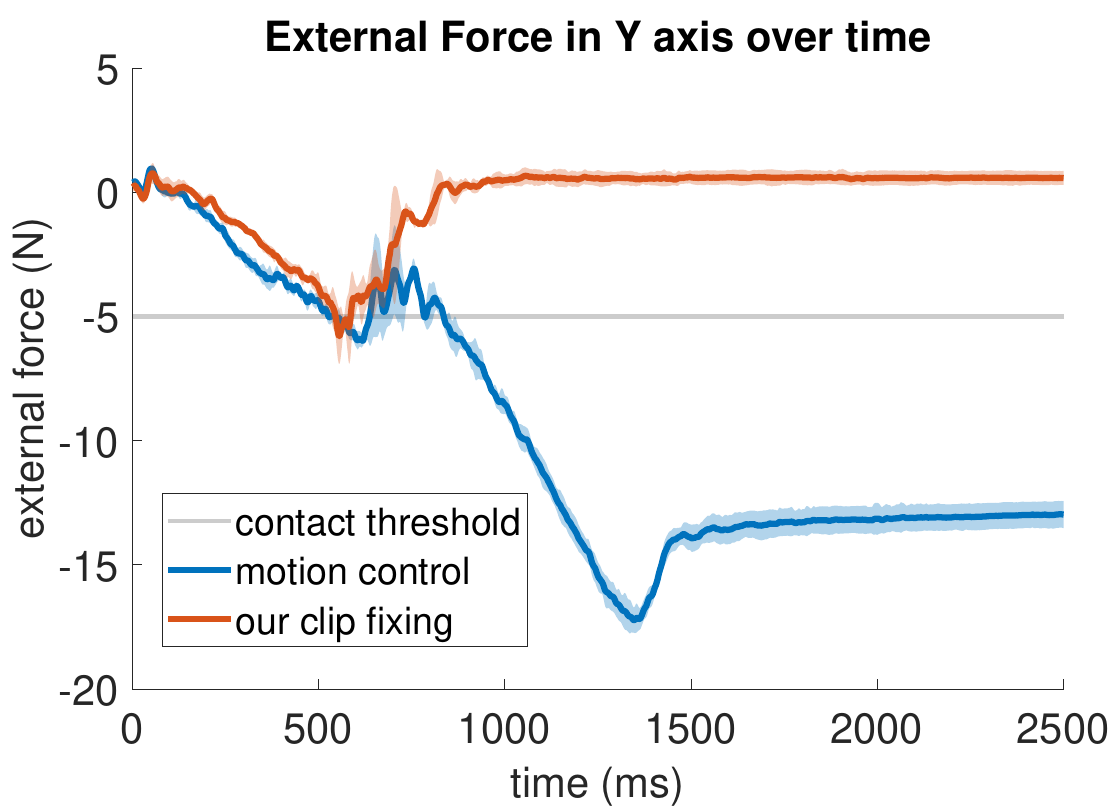}};
        \node[inner sep=0pt] (motion) at (2.1,-4.5) {\includegraphics[width=0.22\textwidth]{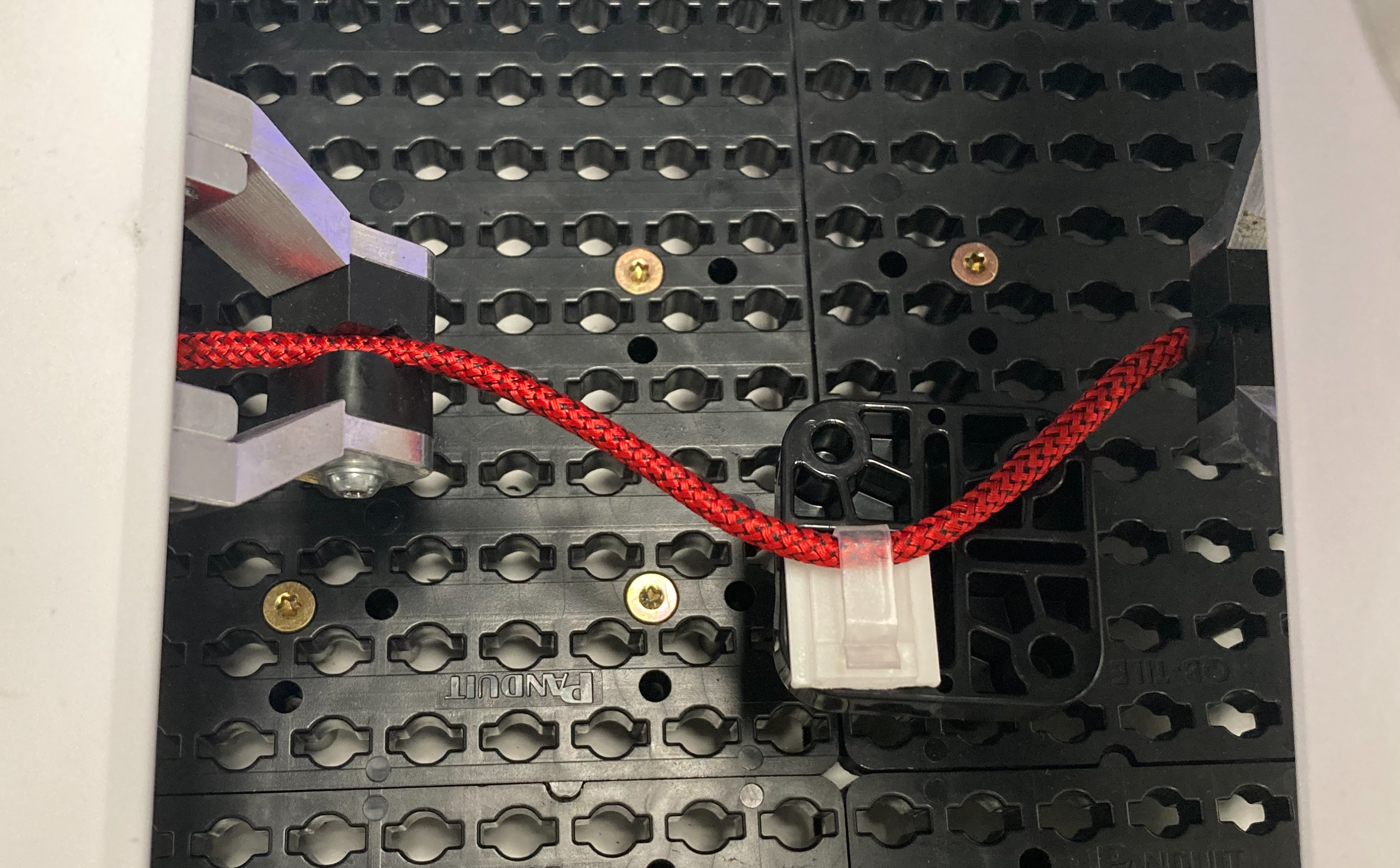}};
        \node[inner sep=0pt] (our) at (-2.1,-4.5) {\includegraphics[width=0.22\textwidth]{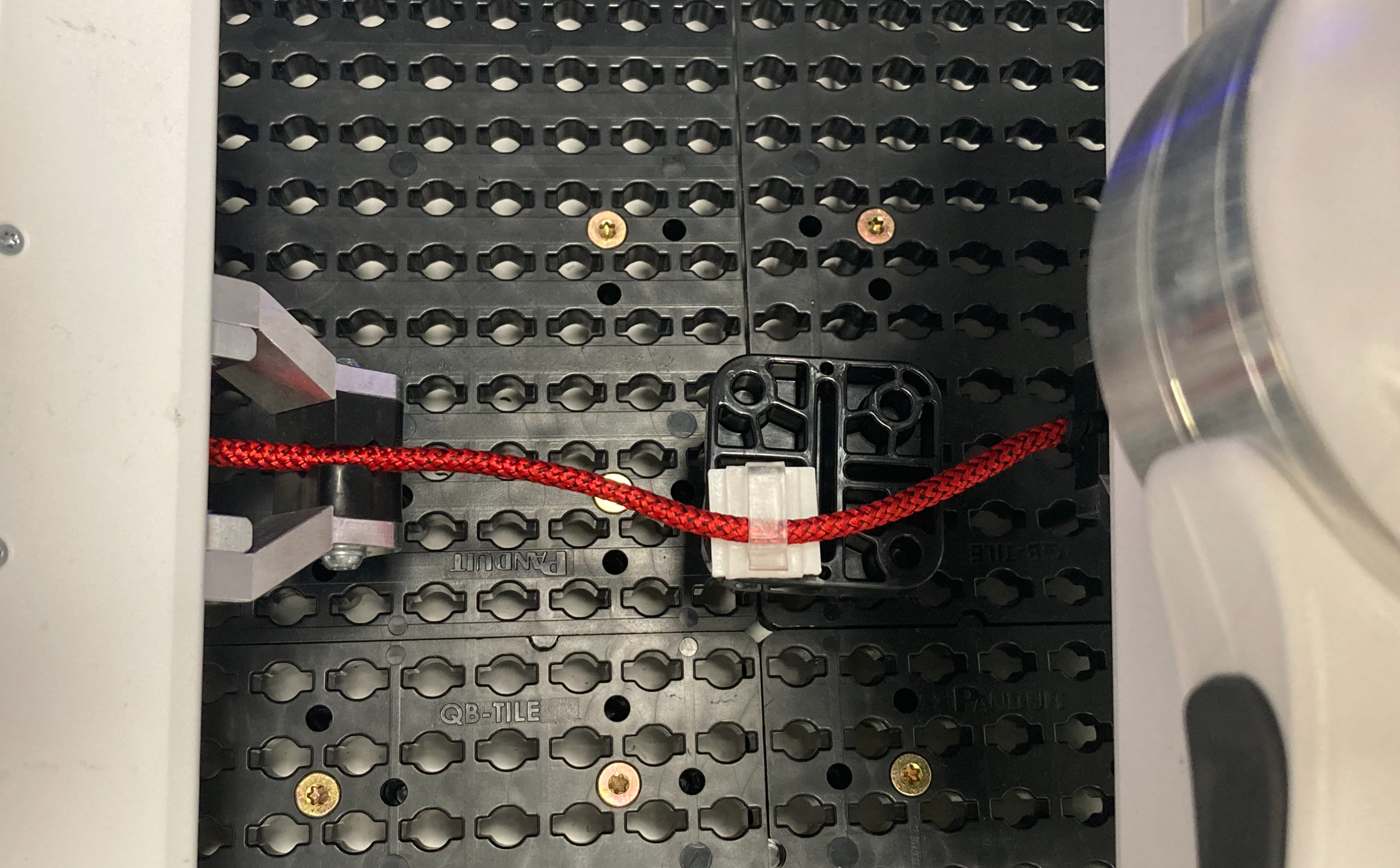}};
    \end{tikzpicture}

    
    \caption{Clip fixing skill evaluation. Top: Contact force in $\mathbf{u}$ direction during clip fixing. Bottom left: The proposed fixing skill results in a final DLO status that is more closely aligned with the clip. Bottom right: The final status of the DLO is overstretched beyond the limit of the clip when using only motion control.}
    \label{fig:exp_clip_fix}
    \vspace{-1.7em}
\end{figure}

\begin{figure*}[!t]
    \centering
    \begin{subfigure}[h]{0.3\textwidth}   
        \centering 
        \includegraphics[width=\textwidth]{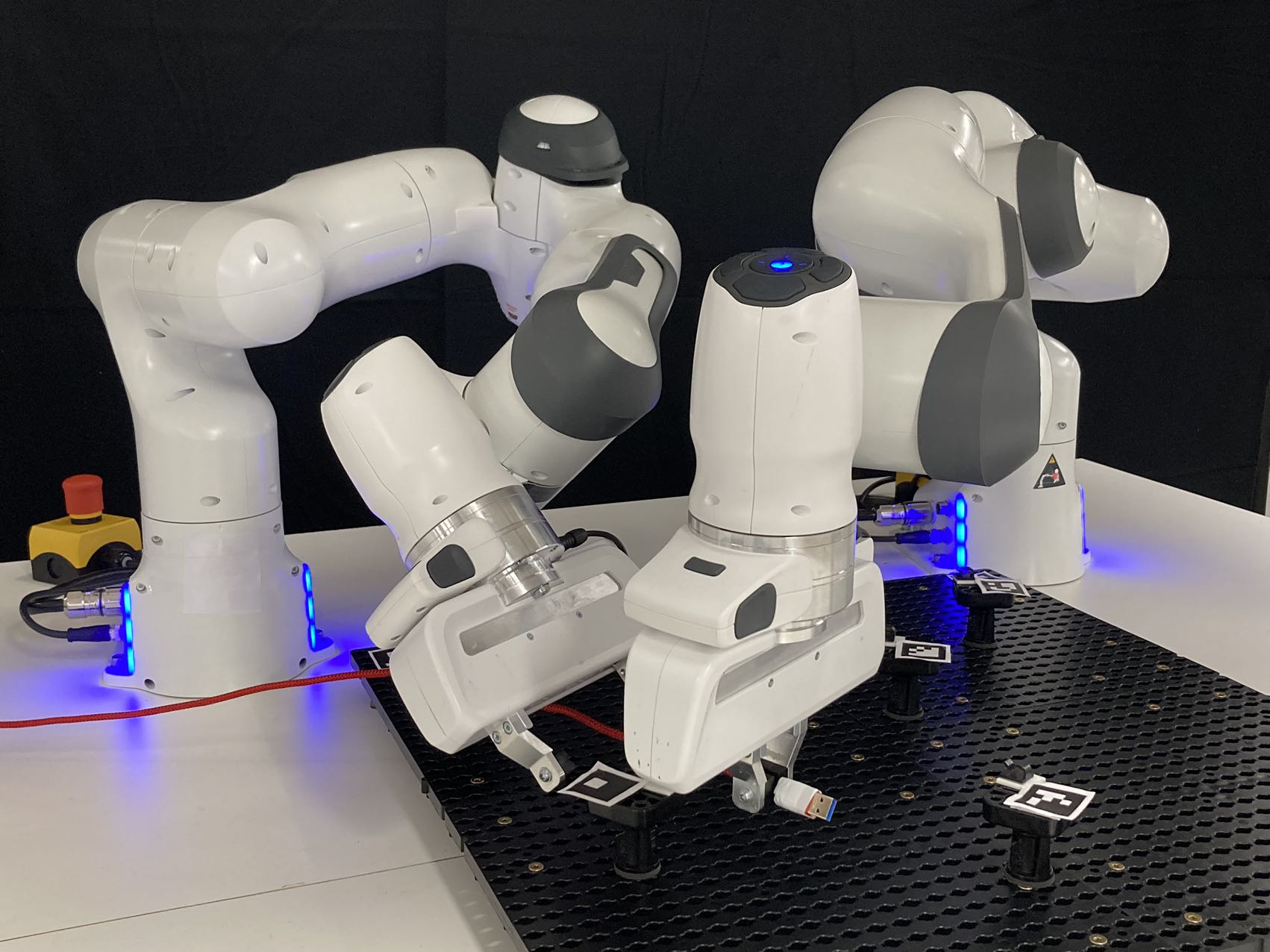}
        \caption{}     
    \end{subfigure}
    \begin{subfigure}[h]{0.3\textwidth}
        \centering
        \includegraphics[width=\textwidth]{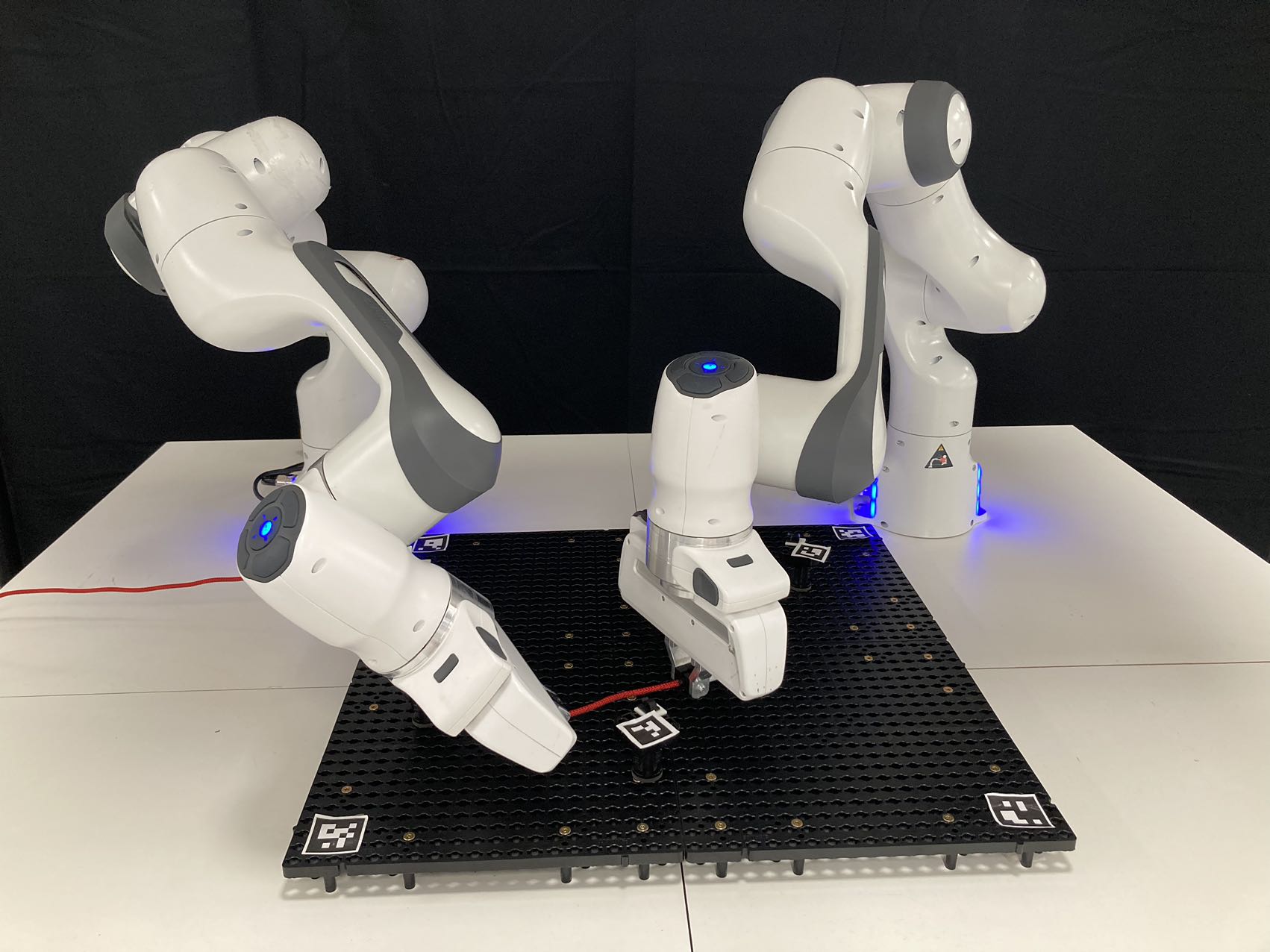}
        \caption{}   
    \end{subfigure}
    \begin{subfigure}[h]{0.3\textwidth}  
        \centering 
        \includegraphics[width=\textwidth]{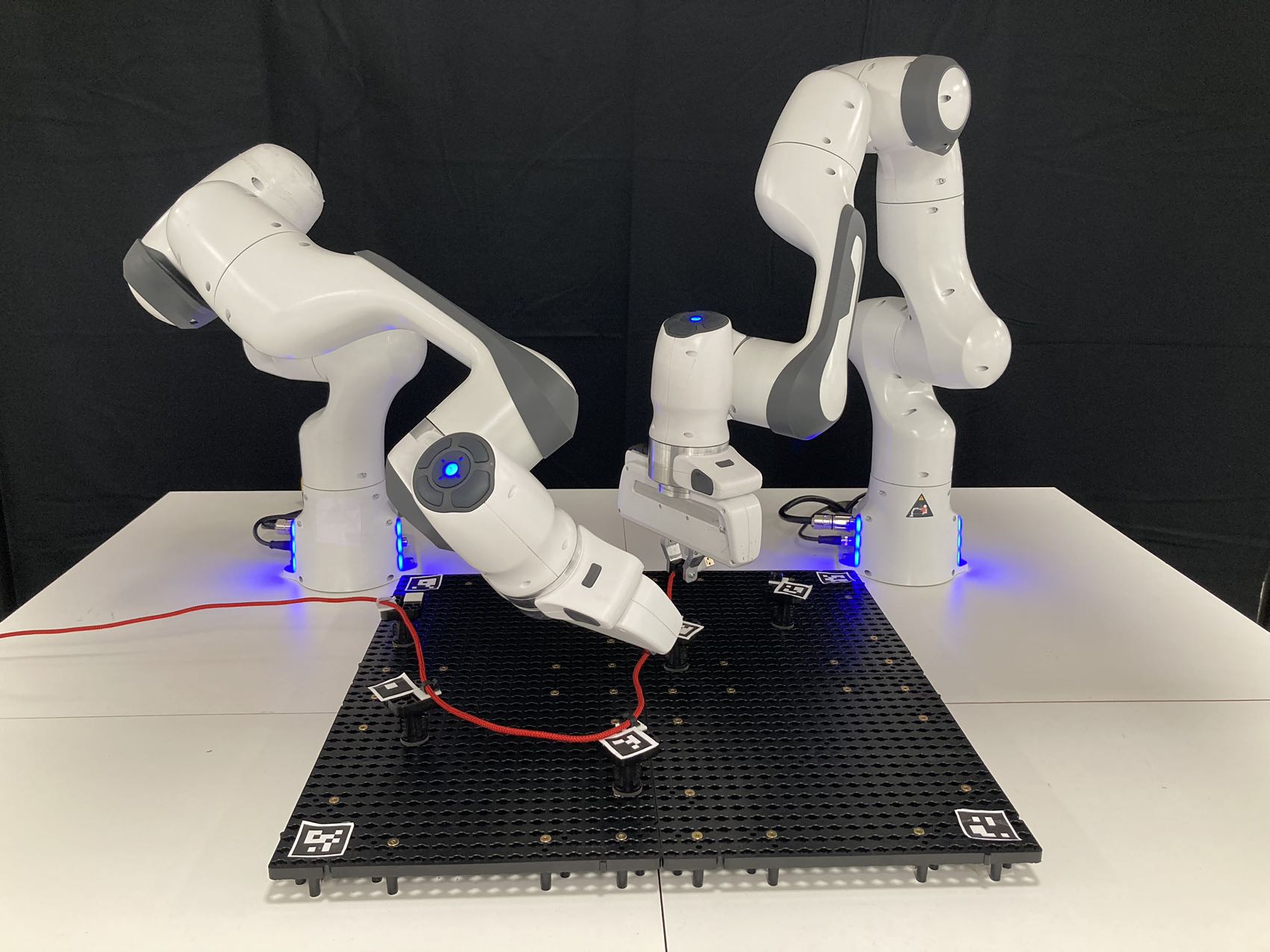}
        \caption{}  
    \end{subfigure}
    \vskip\baselineskip
    \vspace{-1em}
    \begin{subfigure}[h]{0.3\textwidth}   
        \centering 
        \includegraphics[width=\textwidth]{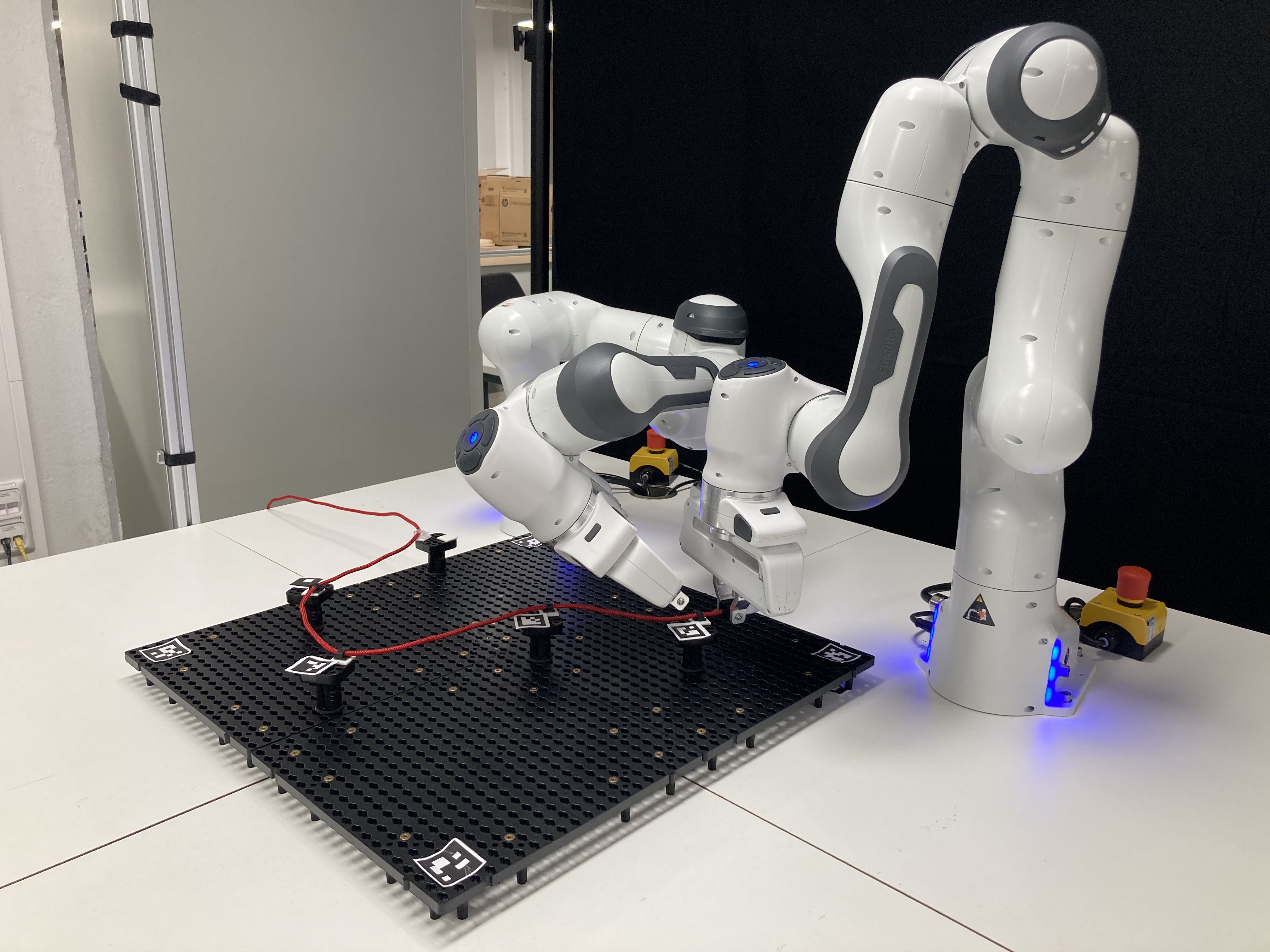}
        \caption{}  
    \end{subfigure}
    \begin{subfigure}[h]{0.3\textwidth}   
        \centering 
        \includegraphics[width=\textwidth, trim={0cm 0 0cm 0},clip]{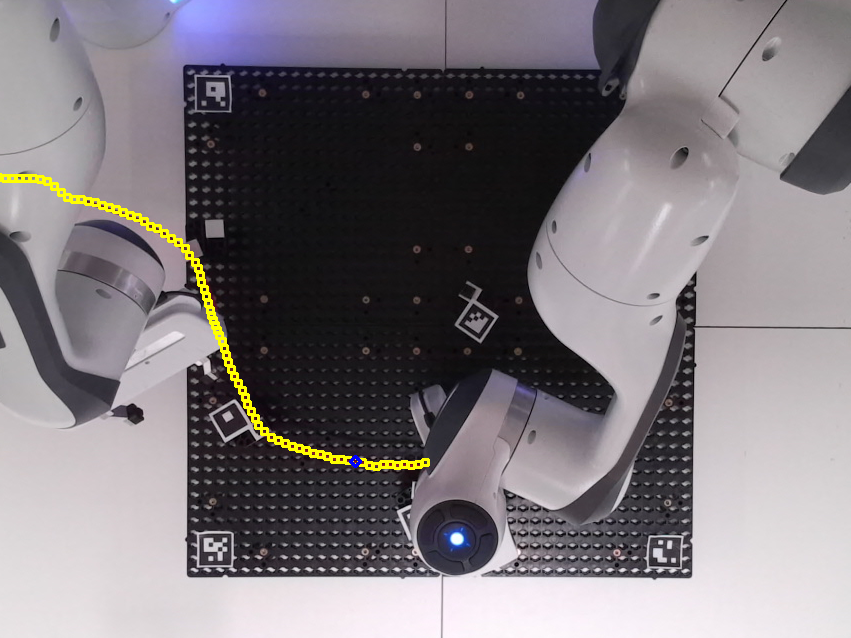}
        \caption{}  
    \end{subfigure}
    \begin{subfigure}[h]{0.3\textwidth}   
        \centering 
        \includegraphics[width=\textwidth, trim={0cm 0cm 0cm 0},clip]{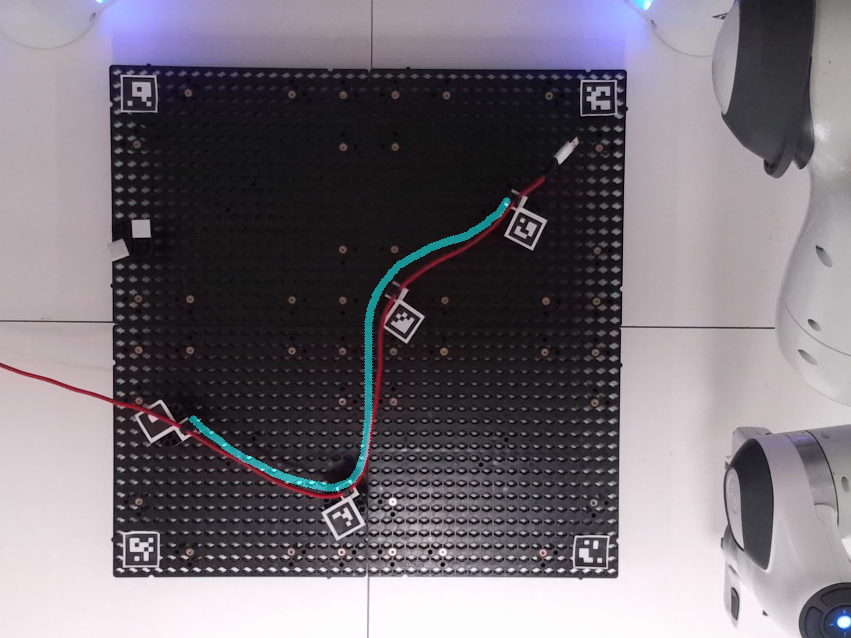}
        \caption{}  
    \end{subfigure}
    \caption{Cable routing process. (a)-(d) DLO manipulation with 4 planned fixtures. (e) Grasping point (blue) from online DLO state estimation (yellow). (f) Achieved shape of DLO (red) compared with desired shape (green). The detailed movement can be viewed in the accompanying video.}
    \label{fig:exp_process}
    \vspace{-1.7em}
\end{figure*}

\subsection{Clip Fixing}


To evaluate the flexibility of our contact-aware clip fixing skill, we compare the detected external force in the fixing process using our skill and using purely motion control. 
We record the external contact force from the robot leader, as it has no other contacts with the environment in addition to the fixture being dealt with. In comparison, the robot follower is also under influence of forces from the last fixture, which can make the contact detection less reliable.

We test both ways of clip fixing on a clip with $2$ cm length in the fixing direction. The contact threshold $F_c$ is set as $5$ N. For the fixing with solely motion control, the movement in $\mathbf{u}$ is set as $4$ cm. 
The detected $f_{\mathrm{ext}}$ is plotted in Fig.~\ref{fig:exp_clip_fix}, where one falling represents one contact with the fixture. Our clip fixing skill experiences only one contact stage, and stops moving further or applying forces once the cable is fitted snugly in (see Fig.~\ref{fig:exp_clip_fix} bottom left). In contrast, the fixing by motion control keeps moving regardless of contact, and hits the back end of the clip (see  Fig.~\ref{fig:exp_clip_fix} bottom right). In the real production process, this blindness may lead to damages. 


\subsection{Cable Routing}
Finally, we evaluate the whole framework with a drawn goal shape as input and a red usb extension cable as object to be manipulated. 
We place fixtures on four positions generated following Algorithm~\ref{alg:place}, all with opening direction $\mathbf{u} = (0, 1, 0)^T$.
To hold the cable on $M_{\mathrm{fixture}}$ initially, we place an assistive fixture at the upper left corner of the harness board (the fixture without marker) which can be easily removed after the routing process.

The motion planning in Section~\ref{subsec:skill_motion} is implemented with MoveIt~\cite{coleman2014reducing} using the Open Motion Planning Library~\cite{sucan2012open}. Motion sequence of both robots passing each placed fixture is depicted in Fig.~\ref{fig:exp_process}(a)-(d). 
After the robot leader guides the DLO to one clip fixture, the robot follower approaches the DLO while avoiding collision with the leader and grasps it. 
Both robots then apply stretching and pushing force to the DLO to insert it into the clip fixture. 
One of the grasping points obtained from online DLO status estimation is shown in Fig.~\ref{fig:exp_process}(e). The estimated status is represented as a sequence of yellow points, and the grasping point is marked as blue.
As contacts with all fixtures are established, the robots release their grasp on the DLO, leaving it securely fixed in place by the clips (see Fig.~\ref{fig:exp_process}(f)).
The whole motion and clip fixing process can be found in the video at \url{https://youtu.be/YbVDOgT3vc4}.




\section{Conclusion}
\vspace{-0.5em}
We presented a framework to manipulate DLO with environmental contacts, with DLO status estimated online from visual and force information. 
Taking a shape as input, our fixture placing algorithm firstly generates appropriate positions of fixtures. 
The DLO status is estimated online with information from visual sensors as well as force-torque sensors. 
An object-centered skill engine then selects from two designed manipulation skills based on current DLO status to establish contact with each fixture. 
Real=world experiment results show that our framework successfully controls two robots to manipulate the DLO into the desired shape. 
The contact-aware clip-fixing skill especially outperforms the solely motion-control based fixing in terms of flexibility.

For future work, we plan to include learning strategies to combine skills and obtain skill parameters in a more adaptive way~\cite{nasiriany2022augmenting,bing2021complex,bing2022solving,bing2022meta}. A more complex shape could be probably achieved with the help of grippers with more degrees of freedom as well as tactile sensory information~\cite{she2021cable,wilson2023cable}.









\bibliographystyle{IEEEtran}
\bibliography{references}

\begin{thebibliography}{10}
\providecommand{\url}[1]{#1}
\csname url@samestyle\endcsname
\providecommand{\newblock}{\relax}
\providecommand{\bibinfo}[2]{#2}
\providecommand{\BIBentrySTDinterwordspacing}{\spaceskip=0pt\relax}
\providecommand{\BIBentryALTinterwordstretchfactor}{4}
\providecommand{\BIBentryALTinterwordspacing}{\spaceskip=\fontdimen2\font plus
\BIBentryALTinterwordstretchfactor\fontdimen3\font minus
  \fontdimen4\font\relax}
\providecommand{\BIBforeignlanguage}[2]{{%
\expandafter\ifx\csname l@#1\endcsname\relax
\typeout{** WARNING: IEEEtran.bst: No hyphenation pattern has been}%
\typeout{** loaded for the language `#1'. Using the pattern for}%
\typeout{** the default language instead.}%
\else
\language=\csname l@#1\endcsname
\fi
#2}}
\providecommand{\BIBdecl}{\relax}
\BIBdecl

\bibitem{she2021cable}
Y.~She, S.~Wang, S.~Dong, N.~Sunil, A.~Rodriguez, and E.~Adelson, ``Cable
  manipulation with a tactile-reactive gripper,'' \emph{The International
  Journal of Robotics Research}, vol.~40, no. 12-14, pp. 1385--1401, 2021.

\bibitem{navarro2017fourier}
D.~Navarro-Alarcon and Y.-H. Liu, ``Fourier-based shape servoing: A new
  feedback method to actively deform soft objects into desired 2-d image
  contours,'' \emph{IEEE Transactions on Robotics}, vol.~34, no.~1, pp.
  272--279, 2017.

\bibitem{zhu2022challenges}
J.~Zhu, A.~Cherubini, C.~Dune, D.~Navarro-Alarcon, F.~Alambeigi, D.~Berenson,
  F.~Ficuciello, K.~Harada, J.~Kober, X.~Li \emph{et~al.}, ``Challenges and
  outlook in robotic manipulation of deformable objects,'' \emph{IEEE Robotics
  \& Automation Magazine}, vol.~29, no.~3, pp. 67--77, 2022.

\bibitem{nadon2018multi}
F.~Nadon, A.~J. Valencia, and P.~Payeur, ``Multi-modal sensing and robotic
  manipulation of non-rigid objects: A survey,'' \emph{Robotics}, vol.~7,
  no.~4, p.~74, 2018.

\bibitem{sanchez2018robotic}
J.~Sanchez, J.-A. Corrales, B.-C. Bouzgarrou, and Y.~Mezouar, ``Robotic
  manipulation and sensing of deformable objects in domestic and industrial
  applications: a survey,'' \emph{The International Journal of Robotics
  Research}, vol.~37, no.~7, pp. 688--716, 2018.

\bibitem{herguedas2019survey}
R.~Herguedas, G.~L{\'o}pez-Nicol{\'a}s, R.~Arag{\"u}{\'e}s, and
  C.~Sag{\"u}{\'e}s, ``Survey on multi-robot manipulation of deformable
  objects,'' in \emph{2019 24th IEEE International Conference on Emerging
  Technologies and Factory Automation (ETFA)}.\hskip 1em plus 0.5em minus
  0.4em\relax IEEE, 2019, pp. 977--984.

\bibitem{jia2018manipulating}
B.~Jia, Z.~Hu, J.~Pan, and D.~Manocha, ``Manipulating highly deformable
  materials using a visual feedback dictionary,'' in \emph{2018 IEEE
  International Conference on Robotics and Automation (ICRA)}.\hskip 1em plus
  0.5em minus 0.4em\relax IEEE, 2018, pp. 239--246.

\bibitem{zhu2018dual}
J.~Zhu, B.~Navarro, P.~Fraisse, A.~Crosnier, and A.~Cherubini, ``Dual-arm
  robotic manipulation of flexible cables,'' in \emph{2018 IEEE/RSJ
  International Conference on Intelligent Robots and Systems (IROS)}.\hskip 1em
  plus 0.5em minus 0.4em\relax IEEE, 2018, pp. 479--484.

\bibitem{kruse2015collaborative}
D.~Kruse, R.~J. Radke, and J.~T. Wen, ``Collaborative human-robot manipulation
  of highly deformable materials,'' in \emph{2015 IEEE international conference
  on robotics and automation (ICRA)}.\hskip 1em plus 0.5em minus 0.4em\relax
  IEEE, 2015, pp. 3782--3787.

\bibitem{mcconachie2020manipulating}
D.~McConachie, A.~Dobson, M.~Ruan, and D.~Berenson, ``Manipulating deformable
  objects by interleaving prediction, planning, and control,'' \emph{The
  International Journal of Robotics Research}, vol.~39, no.~8, pp. 957--982,
  2020.

\bibitem{zhu2019robotic}
J.~Zhu, B.~Navarro, R.~Passama, P.~Fraisse, A.~Crosnier, and A.~Cherubini,
  ``Robotic manipulation planning for shaping deformable linear objects with
  environmental contacts,'' \emph{IEEE Robotics and Automation Letters},
  vol.~5, no.~1, pp. 16--23, 2019.

\bibitem{huo2022keypoint}
S.~Huo, A.~Duan, C.~Li, P.~Zhou, W.~Ma, H.~Wang, and D.~Navarro-Alarcon,
  ``Keypoint-based planar bimanual shaping of deformable linear objects under
  environmental constraints with hierarchical action framework,'' \emph{IEEE
  Robotics and Automation Letters}, vol.~7, no.~2, pp. 5222--5229, 2022.

\bibitem{jin2022robotic}
S.~Jin, W.~Lian, C.~Wang, M.~Tomizuka, and S.~Schaal, ``Robotic cable routing
  with spatial representation,'' \emph{IEEE Robotics and Automation Letters},
  vol.~7, no.~2, pp. 5687--5694, 2022.

\bibitem{lagneau2020automatic}
R.~Lagneau, A.~Krupa, and M.~Marchal, ``Automatic shape control of deformable
  wires based on model-free visual servoing,'' \emph{IEEE Robotics and
  Automation Letters}, vol.~5, no.~4, pp. 5252--5259, 2020.

\bibitem{yu2022shape}
M.~Yu, H.~Zhong, and X.~Li, ``Shape control of deformable linear objects with
  offline and online learning of local linear deformation models,'' in
  \emph{2022 International Conference on Robotics and Automation (ICRA)}.\hskip
  1em plus 0.5em minus 0.4em\relax IEEE, 2022, pp. 1337--1343.

\bibitem{sintov2020motion}
A.~Sintov, S.~Macenski, A.~Borum, and T.~Bretl, ``Motion planning for dual-arm
  manipulation of elastic rods,'' \emph{IEEE Robotics and Automation Letters},
  vol.~5, no.~4, pp. 6065--6072, 2020.

\bibitem{waltersson2022planning}
G.~A. Waltersson, R.~Laezza, and Y.~Karayiannidis, ``Planning and control for
  cable-routing with dual-arm robot,'' in \emph{2022 International Conference
  on Robotics and Automation (ICRA)}.\hskip 1em plus 0.5em minus 0.4em\relax
  IEEE, 2022, pp. 1046--1052.

\bibitem{suberkrub2022feel}
F.~S{\"u}berkr{\"u}b, R.~Laezza, and Y.~Karayiannidis, ``Feel the tension:
  Manipulation of deformable linear objects in environments with fixtures using
  force information,'' in \emph{2022 IEEE/RSJ International Conference on
  Intelligent Robots and Systems (IROS)}.\hskip 1em plus 0.5em minus
  0.4em\relax IEEE, 2022, pp. 11\,216--11\,222.

\bibitem{caporali2022fastdlo}
A.~Caporali, K.~Galassi, R.~Zanella, and G.~Palli, ``Fastdlo: Fast deformable
  linear objects instance segmentation,'' \emph{IEEE Robotics and Automation
  Letters}, vol.~7, no.~4, pp. 9075--9082, 2022.

\bibitem{Yang2011HumanLike}
C.~Yang, G.~Ganesh, S.~Haddadin, S.~Parusel, A.~Albu-Schaeffer, and E.~Burdet,
  ``Human-like adaptation of force and impedance in stable and unstable
  interactions,'' vol.~27, no.~5, pp. 918--930.

\bibitem{johannsmeier2019framework}
L.~Johannsmeier, M.~Gerchow, and S.~Haddadin, ``A framework for robot
  manipulation: Skill formalism, meta learning and adaptive control,'' in
  \emph{2019 International Conference on Robotics and Automation (ICRA)}.\hskip
  1em plus 0.5em minus 0.4em\relax IEEE, 2019, pp. 5844--5850.

\bibitem{coleman2014reducing}
D.~Coleman, I.~Sucan, S.~Chitta, and N.~Correll, ``Reducing the barrier to
  entry of complex robotic software: a moveit! case study,'' \emph{arXiv
  preprint arXiv:1404.3785}, 2014.

\bibitem{sucan2012open}
I.~A. Sucan, M.~Moll, and L.~E. Kavraki, ``The open motion planning library,''
  \emph{IEEE Robotics \& Automation Magazine}, vol.~19, no.~4, pp. 72--82,
  2012.

\bibitem{nasiriany2022augmenting}
S.~Nasiriany, H.~Liu, and Y.~Zhu, ``Augmenting reinforcement learning with
  behavior primitives for diverse manipulation tasks,'' in \emph{2022
  International Conference on Robotics and Automation (ICRA)}.\hskip 1em plus
  0.5em minus 0.4em\relax IEEE, 2022, pp. 7477--7484.

\bibitem{bing2021complex}
Z.~Bing, M.~Brucker, F.~O. Morin, R.~Li, X.~Su, K.~Huang, and A.~Knoll,
  ``Complex robotic manipulation via graph-based hindsight goal generation,''
  \emph{IEEE transactions on neural networks and learning systems}, vol.~33,
  no.~12, pp. 7863--7876, 2021.

\bibitem{bing2022solving}
Z.~Bing, H.~Zhou, R.~Li, X.~Su, F.~O. Morin, K.~Huang, and A.~Knoll, ``Solving
  robotic manipulation with sparse reward reinforcement learning via
  graph-based diversity and proximity,'' \emph{IEEE Transactions on Industrial
  Electronics}, vol.~70, no.~3, pp. 2759--2769, 2022.

\bibitem{bing2022meta}
Z.~Bing, D.~Lerch, K.~Huang, and A.~Knoll, ``Meta-reinforcement learning in
  non-stationary and dynamic environments,'' \emph{IEEE Transactions on Pattern
  Analysis and Machine Intelligence}, vol.~45, no.~3, pp. 3476--3491, 2022.

\bibitem{wilson2023cable}
A.~Wilson, H.~Jiang, W.~Lian, and W.~Yuan, ``Cable routing and assembly using
  tactile-driven motion primitives,'' \emph{arXiv preprint arXiv:2303.11765},
  2023.

\end{thebibliography}

\balance

\end{document}